\documentclass[runningheads]{llncs}
\usepackage[T1]{fontenc}
\usepackage[utf8]{inputenc}
\title{Enhanced Kalman with Adaptive Appearance Motion SORT for Grounded Generic Multiple Object Tracking}
\titlerunning{G2MOT}
\author{Duy Le Dinh Anh\inst{1} \and
Kim Hoang Tran\inst{1} \and
Quang-Thuc Nguyen\inst{2, 3} \and
Ngan Hoang Le\inst{1}}
\authorrunning{Duy et al.}
\institute{University of Arkansas, AR, USA \and
Faculty of Information Technology and Software Engineering Lab, University of Science, VNU-HCM \and
Vietnam National University, Ho Chi Minh City, Vietnam}

\usepackage{graphicx}
\usepackage{booktabs}
\usepackage[toc,page]{appendix}

\usepackage[accsupp]{axessibility}  

\usepackage{hyperref}
\usepackage[accsupp]{axessibility}  

\usepackage{times}
\usepackage{latexsym}

\usepackage[T1]{fontenc}

\usepackage[utf8]{inputenc}

\usepackage{microtype}

\usepackage{inconsolata}

\usepackage{soul}
\usepackage{cleveref}
\usepackage[normalem]{ulem}
\usepackage{algorithmic}
\usepackage{algorithm2e}
\usepackage{xcolor}
\usepackage{graphicx}
\usepackage{hyperref}
\usepackage{amsmath}
\usepackage{ragged2e}
\usepackage{blindtext}
\usepackage{amsfonts}
\usepackage{multirow}
\usepackage{color}
\usepackage{booktabs, cellspace, multirow, tabularx}
\usepackage{array}
\usepackage{listings}
\usepackage[T1]{fontenc}
\usepackage{colortbl}
\usepackage{wrapfig,lipsum,booktabs}
\usepackage{amssymb}
\usepackage{pifont}
\usepackage{subcaption}
\usepackage{enumitem}
\usepackage{lipsum}

\usepackage{amssymb}
\usepackage{pifont}
\newcommand{\cmark}{\ding{51}}%
\newcommand{\xmark}{\ding{55}}%

\usepackage{orcidlink}
\usepackage{floatflt}
\usepackage{subcaption}

\definecolor{airforceblue}{rgb}{0.36, 0.54, 0.66}
\definecolor{asparagus}{rgb}{0.94, 0.87, 0.8}
\definecolor{almond}{rgb}{0.53, 0.66, 0.42}
\definecolor{babyblue}{rgb}{0.54, 0.81, 0.94}
\definecolor{aureolin}{rgb}{0.99, 0.93, 0.0}

\definecolor{classname}{rgb}{0.98, 0.92, 0.84}
\definecolor{type}{rgb}{0.7, 0.75, 0.71}
\definecolor{synonym}{rgb}{0.96, 0.76, 0.76}
\definecolor{caption}{rgb}{0.54, 0.81, 0.94}
\definecolor{definition}{rgb}{0.67, 0.9, 0.93}
\definecolor{attribute}{rgb}{0.96, 0.73, 1.0}
\definecolor{track_path}{rgb}{1.0, 0.33, 0.64}

\definecolor{airforceblue}{rgb}{0.36, 0.54, 0.66}
\definecolor{asparagus}{rgb}{0.94, 0.87, 0.8}
\definecolor{almond}{rgb}{0.53, 0.66, 0.42}
\definecolor{babyblue}{rgb}{0.54, 0.81, 0.94}
\definecolor{aureolin}{rgb}{0.99, 0.93, 0.0}

\definecolor{classname}{rgb}{0.98, 0.92, 0.84}
\definecolor{type}{rgb}{0.7, 0.75, 0.71}
\definecolor{synonym}{rgb}{0.96, 0.76, 0.76}
\definecolor{caption}{rgb}{0.54, 0.81, 0.94}
\definecolor{definition}{rgb}{0.67, 0.9, 0.93}
\definecolor{attribute}{rgb}{0.96, 0.73, 1.0}

\begin{document}

\title{Enhanced Kalman with Adaptive Appearance Motion SORT for Grounded Generic Multiple Object Tracking} 

\maketitle

\begin{abstract}
Despite recent progress, Multi-Object Tracking (MOT) continues to face significant challenges, particularly its dependence on prior knowledge and predefined categories, complicating the tracking of unfamiliar objects. Generic Multiple Object Tracking (GMOT) emerges as a promising solution, requiring less prior information. Nevertheless, existing GMOT methods, primarily designed as OneShot-GMOT, rely heavily on initial bounding boxes and often struggle with variations in viewpoint, lighting, occlusion, and scale. To overcome the limitations inherent in both MOT and GMOT when it comes to tracking objects with specific generic attributes, we introduce \textbf{Grounded-GMOT}, an innovative tracking paradigm that enables users to track multiple generic objects in videos through natural language descriptors.
Our contributions begin with the introduction of the \textbf{G$^2$MOT dataset}, which includes a collection of videos featuring a wide variety of generic objects, each accompanied by detailed textual descriptions of their attributes. Following this, we propose a novel tracking method, \textbf{KAM-SORT}, which not only effectively integrates visual appearance with motion cues but also enhances the Kalman filter. KAM-SORT proves particularly advantageous when dealing with objects of high visual similarity from the same generic category in GMOT scenarios. Through comprehensive experiments, we demonstrate that Grounded-GMOT outperforms existing OneShot-GMOT approaches. Additionally, our extensive comparisons between various trackers highlight KAM-SORT's efficacy in GMOT, further establishing its significance in the field. Project page: \url{https://UARK-AICV.github.io/G2MOT}. The source code and dataset will be made publicly available.

\keywords{Generic MOT \and Grounded GMOT \and G2MOT \and KAM-SORT}
\end{abstract}
\vspace{-2em}
\section{Introduction}
\label{sec:intro}

\begin{figure}
    \centering
    \includegraphics[width=\linewidth]{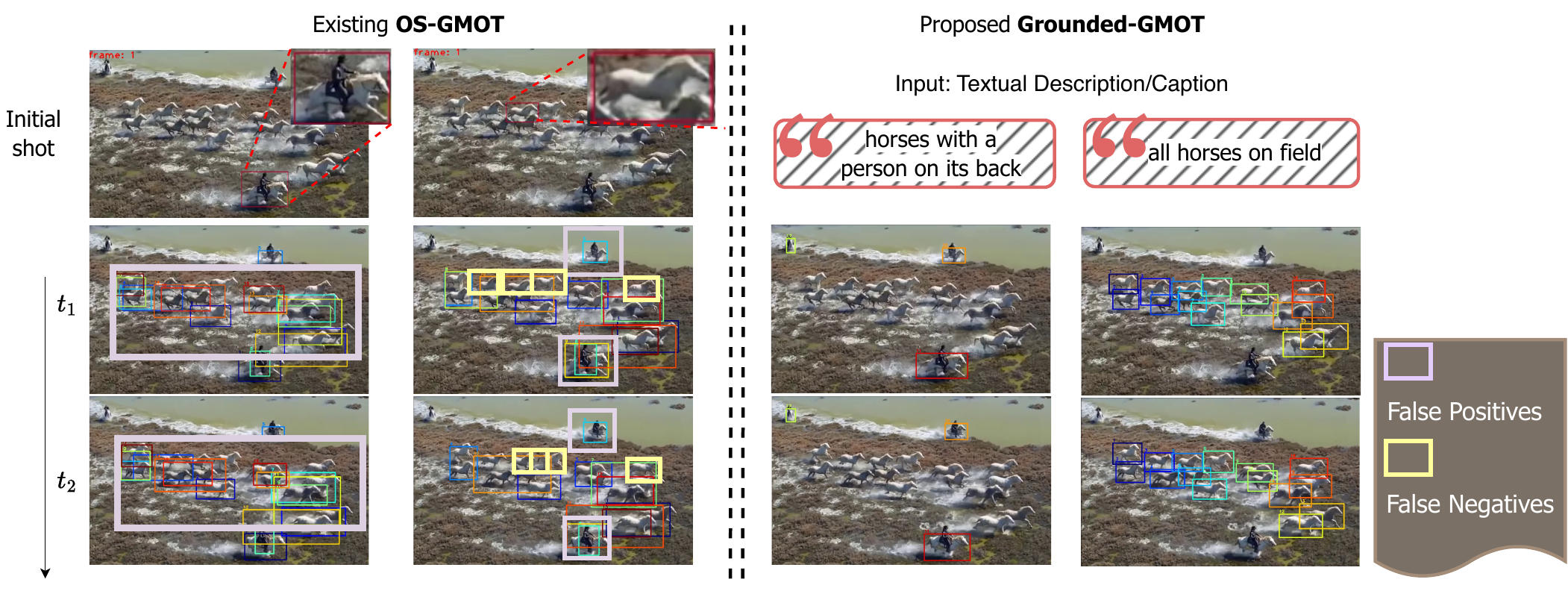}
    \caption{Comparison between OneShot-GMOT (OS-GMOT) (left) and our Grounded-GMOT (right) in tracking multiple generic objects. The tracking system receives input ($1^{st}$ row): OS-GMOT relies on an initial bounding box, and Grounded-GMOT utilizes textual descriptions. OS-GMOT encounters numerous challenges related to pose, illumination, occlusion, scale, texture, etc, resulting in many False Positives and False Negatives corresponding to two different initial bounding boxes. Our Grounded-GMOT adeptly detects objects based on input queries and tracks them over time at intervals $t_1$ and $t_2$ ($2^{nd}$ and $3^{rd}$ rows). \vspace{-5mm}}
    \label{fig:teaser}
\end{figure}

Multiple Object Tracking (MOT) \cite{bewley2016simple,leal2016learning,wojke2017simple,braso2020learning,wu2021track,cao2023observation,maggiolino2023deep,zhang2021bytetrack,yan2022towards,meinhardt2022trackformer,zeng2022motr,cai2022memot} plays a crucial role in dynamic scene analysis, proving essential for various critical real-world applications including surveillance, security, autonomous driving, robotics, and biology. Despite significant advancements in this field, current MOT methodologies predominantly focus on a limited set of object categories, typically emphasizing a specific area of interest, such as pedestrians \cite{MOTChallenge2015,MOT16,dendorfer2020mot20}, or objects pertinent to autonomous driving scenarios \cite{chen2018bdd100k,caesar2020nuscenes}. These approaches require a substantial amount of prior knowledge about the objects being tracked and depend heavily on large, extensively labeled datasets. As a result, they face challenges in tracking objects across unseen or specific categories, as well as in handling objects with indistinguishable features.

Generic Multiple Object Tracking (GMOT) \cite{luo2013generic,luo2014bi,bai2021gmot} aims to alleviate these limitations by reducing the dependency on prior information. GMOT is designed to track multiple objects of a common or similar generic type, making it suitable for a wide array of applications, ranging from annotation and video editing to monitoring animal behavior. Notwithstanding, conventional GMOT methodologies \cite{luo2013generic,luo2014bi,bai2021gmot} are predominantly anchored in a one-shot paradigm, i.e. OneShot-GMOT (OS-GMOT), leveraging the initial bounding box of a single target object in the first frame to track all objects of the same class. 
The dependency on the starting bounding box poses challenges in accommodating object variations e.g., pose, illumination, occlusion, scale, texture, etc.

In recent years, significant strides have been made in achieving grounded understanding through the integration of natural language processing into computer vision \cite{liu2019aligning,zhang2021consensus,li2023gligen,li2022grounded}. This progress has enabled the precise alignment of language concepts with visual observations, allowing for a comprehensive understanding of both visual content and the nuance of natural language. Building upon this foundation, we aim to address the limitations of both MOT and GMOT in tracking objects with specific generic attributes. To this end, we introduce a novel tracking paradigm called \textbf{Grounded-GMOT}, which leverages the capabilities of Vision Language Models (VLMs) to guide the tracking of multiple generic objects in videos using descriptive natural language input. Figure \ref{fig:teaser} shows the comparison between OS-GMOT and our proposed Grounded-GMOT. 

In this work, we first introduce \textbf{G$^2$MOT dataset}, a large-scale dataset enriched with a variety of generic object categories and their corresponding textual descriptions. G$^2$MOT dataset surpasses all currently available datasets in terms of size and diversity, as in Table \ref{tb:dataset_comparison}. 
We then propose \textbf{KAM-SORT} (Enhanced \underline{\textbf{K}}alman Filter with Adaptive \underline{\textbf{A}}ppearance \underline{\textbf{M}}otion \underline{\textbf{SORT}}), an innovative GMOT tracker. Our KAM-SORT tracker first enhances the Kalman Filter by integrating camera motion into the re-association of predicted bounding boxes. Subsequently, it adeptly measures appearance uniformity and dynamically adjusts the weighting between motion and appearance in the association process, ensuring robust and accurate object tracking. Our contributions are summarized as follows:

\noindent
$\bullet$ Introducing Grounded-GMOT, a novel tracking paradigm that utilizes responsive and interactive natural language descriptions to track generic objects in videos.

\noindent
$\bullet$ Unveiling the G$^2$MOT dataset, a novel, large-scale dataset encompassing a broad diversity of generic object categories and detailed natural language descriptions.

\noindent
$\bullet$ Providing a comprehensive comparison between SOTA OS-GMOT and Grounded-GMOT, using a variety of popular trackers across numerous performance metrics.

\noindent
$\bullet$ Proposing KAM-SORT, an innovative object association method that enhances the Kalman Filter and adeptly integrates both motion and appearance during the tracking.

\begin{table}[!h]
\vspace{-5mm}
\caption{Comparison of \textbf{existing datasets} of \colorbox{airforceblue!20}{SOT}, \colorbox{almond!50}{MOT}, \colorbox{babyblue!50}{GSOT}, \colorbox{aureolin!50}{GMOT}. ``\textbf{\#}" represents the quantity of the respective items. \textbf{Cat.}, \textbf{Vid.} denote Categories and Videos. \textbf{Obj.}: average number of objects per frame. \textbf{App.}: appearance similarity (\%) between objects in a frame, calculated by the average cosine similarity of objects in the same frame; \textbf{Den.} density of objects in a frame, computed by the maximum number of objects at the same pixel. \textbf{Occ.}: occlusion between objects in a frame, represented by the average ratio of IoU (\%) of the bounding boxes in the same frame; \textbf{Mot.}: motion speed of objects in a video, calculated by the average ratio of the IoU (\%) of the bounding boxes in the same track in consecutive frames.} 
\setlength{\tabcolsep}{3pt}
\renewcommand{\arraystretch}{1.0}
\resizebox{\linewidth}{!}{
\begin{tabular}{l|lc|lllll|ccccc}
\toprule
\multirow{2}{*}{\textbf{Datasets}}   & \multirow{2}{*}{\textbf{Task}} & \multirow{2}{*}{\textbf{NLP}} & \multicolumn{5}{c|}{\textbf{Statistical Information}}& \multicolumn{5}{c}{\textbf{Data Properties (mean(std))}} \\ \cline{4-13}
& & & \#\textbf{Cat.} & \#\textbf{Vid.} & \#\textbf{Frames} & \#\textbf{Tracks} & \#\textbf{Boxs} & \textbf{Obj.} & \textbf{App.} & \textbf{Den.} & \textbf{Occ.} & \textbf{Mot.}\\ \midrule
\rowcolor{airforceblue!20}
OTB2013~\cite{wu2013online} & SOT  &  \xmark   &       10       &    51      &      29K    &      51    &    29K  & -- & -- & -- & -- & --\\ 
\rowcolor{airforceblue!20}
VOT2017~\cite{kristan2016novel} & SOT  &  \xmark   &      24         &    60      &     21K      &     60     &   21K  & -- &  -- & -- & --&  --\\ 
\rowcolor{airforceblue!20}
TrackingNet~\cite{muller2018trackingnet} & SOT  &  \xmark   &       21       &    31K      &   14M       &   31K       &    14M &  -- & -- & -- & -- &  -- \\ 
\rowcolor{almond!50}
MOT17~\cite{milan2016mot16}       & MOT  &  \xmark   &      1        &    14      &    11.2K      &    1.3K      &   0.3M  & 39(35)  & 62(10) & 3.85(1.50) & 14(16) & 94(11)\\ 
\rowcolor{almond!50}
MOT20~\cite{dendorfer2020mot20}        & MOT  &  \xmark    &        1      &     8     &     13.41K     &     3.45K     &     1.65M  & 150(70) & 68(8) & 6.42(1.20) & 15(15) & 96(4) \\ 
\rowcolor{almond!50}
Omni-MOT~\cite{sun2020simultaneous}     & MOT  &  \xmark    &          1    &     --     &      14M+    &    250K      &    110M  & -- & -- & --  & -- & -- \\  
\rowcolor{almond!50}
DanceTrack~\cite{sun2022dancetrack}   & MOT  &  \xmark    &     1         &    100      &     105K     &    990      &  --   & 9(5)  & 77(7) &  2.67(0.99) & 21(17) & 90(9)  \\  
\rowcolor{almond!50}
TAO~\cite{dave2020tao}     & MOT  &  \xmark    &       833       &     2.9K     &    2.6M      &    17.2K      &    333K  & 3(2) & 69(7) & 1.82(0.76)  & 11(14) & 49(34) \\
\rowcolor{almond!50}
SportsMOT~\cite{cui2023sportsmot}     & MOT  &  \xmark    &       1       &    240      &    150K      &     3.4K     &    1.62M  & 11(3)  & 73(8) & 2.44(0.80) & 18(17) & 80(16) \\ 
\rowcolor{babyblue!50}
GOT-10~\cite{huang2019got}      & GSOT  &  \xmark    &      563        &    10K      &    1.5M      &     10K      &    1.5M  & --  & -- & -- & -- & -- \\ 
\rowcolor{babyblue!50}
Fish~\cite{kay2022caltech}         & GSOT  &  \xmark    &      1        &   1.6K       &  527.2K        &     8.25k      &    516K  & -- & -- & --  & -- &  --\\ 
\rowcolor{aureolin!50}
AnimalTrack~\cite{zhang2022animaltrack}  & GMOT &  \xmark   &      10        &     58     &   24.7K       &   1.92K       &    429K  & 17(9)  & 72(8)  & 3.13(1.22) & 15(15) & 91(11) \\ 
\rowcolor{aureolin!50}
GMOT-40~\cite{bai2021gmot}      & GMOT &  \xmark   &        10      &    40      &    9K      &   2.02K       &  256K    & 24(17)  & 71(9) & 2.56(0.88) & 11(12) & 43(44) \\ \midrule
\rowcolor{airforceblue!20}
LaSOT~\cite{fan2019lasot}        &  SOT    &  {coarse}    &      70        &    1.4K      &     3.52M     &    1.4K      &     3.52M  & --  & -- & -- & -- & -- \\ 
\rowcolor{airforceblue!20}
TNL2K~\cite{wang2021towards}        &  SOT    &  {coarse}    &         --     &    2K      &      1.24M    &    2K      &   1.24M   & --  & -- & -- & -- &  -- \\ 
\rowcolor{almond!50}
Refer-KITTI~\cite{wu2023referring}  &  MOT    &  {coarse}   &     2         &     18     &    6.65K      &   637	& 28.72K   & 5(4) & 65(6) & 1.78(0.74)  & 11(11) & 73(21) \\ 
\rowcolor{aureolin!50}
\textbf{G$^2$MOT (Ours)}        & GMOT &   {fine}   &     20         &   253       &   157.2K       &   5.84K    &  1.87M  & 12(5)  &  74(8) & 2.65(0.95) & 18(16) & 84(14) \\  \bottomrule
\end{tabular}}
\vspace{-10mm}
\label{tb:dataset_comparison}
\end{table}

\vspace{-0.25em}
\section{Related Work}
\vspace{-0.25em}
\subsection{Benchmarks}
\vspace{-0.25em}
Recently, numerous benchmark datasets typically fall into two main categories: Varietal Object Tracking (VOT) ~\cite{wu2013online,kristan2016novel,muller2018trackingnet,milan2016mot16,dendorfer2020mot20,sun2020simultaneous,sun2022dancetrack,dave2020tao,cui2023sportsmot,fan2019lasot,wang2021towards,wu2023referring} and Generic Object Tracking (GOT)~\cite{huang2019got,kay2022caltech,zhang2022animaltrack,bai2021gmot}. In VOT, the objects to be tracked typically exhibit diverse visual appearances, while in GOT, the objects share similar visual characteristics. The first category focuses on tracking a single object, encompassing Single Object Tracking (SOT) and Generic Single Object Tracking (GSOT). The second category is dedicated to tracking multiple objects, which includes Multiple Object Tracking (MOT) and Generic Multiple Object Tracking (GMOT). Table \ref{tb:dataset_comparison} shows a detailed comparison of benchmark datasets and their corresponding characteristics.

Most traditional visual tracking datasets \cite{wu2013online,kristan2016novel,muller2018trackingnet,milan2016mot16,dendorfer2020mot20,sun2020simultaneous,sun2022dancetrack,dave2020tao,cui2023sportsmot,huang2019got,kay2022caltech,zhang2022animaltrack,bai2021gmot} have commonly associated labels ID with individual bounding boxes. In contrast, recent tracking datasets \cite{fan2019lasot,wang2021towards,wu2023referring} have incorporated language-assisted captions by harnessing the power of VLMs. However, existing datasets that integrate natural language processing (NLP) with textual descriptions are limited to only SOT and MOT. Our G$^2$MOT dataset goes beyond these limitations, supporting GMOT with rich textual descriptions and offering a significantly larger number of generic object categories and greater diversity in dataset size.

\vspace{-3mm}
\subsection{Pre-trained Vision-Language (VL) models}
\vspace{-2mm}
\label{sec:VLM}
Recent advancements in computer vision have leveraged VL supervision, significantly enhancing model versatility and open-set recognition. A pioneering work in this domain is CLIP \cite{radford2021learning}, which learns visual representations from vast amounts of image-text pairs. Since its release, CLIP has garnered attention, leading to the emergence of several VL models such as ALIGN \cite{jia2021scaling}, ViLD \cite{gu2022open}, RegionCLIP \cite{zhong2022regionclip}, GLIP \cite{li2022grounded,zhang2022glipv2}, Grounding DINO \cite{liu2023grounding}, UniCL \cite{yang2022unified}, X-DETR \cite{cai2022x}, OWL-ViT \cite{minderer2022simple}, LSeg \cite{li2022language}, DenseCLIP \cite{rao2022denseclip}, OpenSeg \cite{ghiasi2022open}, and MaskCLIP \cite{ding2022open}, marking a paradigm shift across vision tasks. VL pre-training models can be categorized into three groups: (i) Image classification: Models like CLIP, ALIGN, and UniCL focus on matching images with language descriptions via bidirectional supervised contrastive learning. (ii) Object detection: This group includes ViLD, RegionCLIP, GLIPv2, X-DETR, OWL-ViT, and Grounding DINO, addressing object localization and recognition. (iii) Image segmentation: The third group involves pixel-level classification using VL models like LSeg, OpenSeg, and DenseSeg. In this work, we employ Grounding DINO as our pre-trained VL model.

\vspace{-3mm} \subsection{Multiple Object Tracking (MOT)} \vspace{-2mm} Object tracking can be broadly categorized into Varietal Object Tracking, including Single Object Tracking (SOT) and MOT, and Generic Object Tracking, comprising Generic Single Object Tracking (GSOT) and GMOT. Our primary focus is on tracking multiple generic objects.

In MOT, approaches are divided based on whether detection and association are executed by a single model or separate models, known as "joint detection and tracking" and "tracking-by-detection." The first category \cite{CHAN2022108793,zhou2020tracking,pang2021quasi,wu2021track,yan2022towards,meinhardt2022trackformer,zeng2022motr} integrates detection into a single network, often with re-identification features. The second category \cite{bewley2016simple,wojke2017simple,zhang2021bytetrack,cao2023observation,maggiolino2023deep} involves a two-step process: detection followed by association with previous tracklets. Tracking-by-detection has achieved SOTA results in MOT, as seen in recent studies like OC-SORT\cite{cao2023observation} and Deep-OCSORT\cite{maggiolino2023deep}.

Despite recent advancements, MOT remains tied to supervised learning and predefined categories, complicating tracking of unfamiliar objects. Unlike MOT, GMOT tracks multiple generic objects without training data, employing a one-shot detection approach called OS-GMOT \cite{bai2021gmot}. While OS-GMOT requires less prior information, it heavily relies on initial bounding boxes and struggles with variations in viewpoint, lighting, occlusion, and scale. In contrast, we introduce a novel zero-shot tracking paradigm, \textit{Grounded-GMOT}, enabling users to track multiple generic objects in videos using natural language descriptors without prior training data or predefined categories.

\vspace{-5mm}
\begin{table}[]
\caption{Statistical information of G$^2$MOT dataset.}
\setlength{\tabcolsep}{3pt}
\renewcommand{\arraystretch}{1.0}
\resizebox{\linewidth}{!}{
\begin{tabular}{l|l|lllll|ccccc}
\toprule
  \multirow{2}{*}{\textbf{Datasets}}         &  \multirow{2}{*}{\textbf{Splits}} & \multicolumn{5}{c|}{\textbf{Statistical Information}}    & \multicolumn{5}{c}{\textbf{Data Properties (mean(std))}} \\   \cline{3-12}
 && \#\textbf{Cat.} & \#\textbf{Vid.} & \#\textbf{Frames} & \#\textbf{Tracks} & \#\textbf{Boxes} & \multicolumn{1}{c}{\textbf{Obj.}} & \multicolumn{1}{c}{\textbf{Mot.}} & \multicolumn{1}{c}{\textbf{Occ.}} & \multicolumn{1}{c}{\textbf{App.}} & \multicolumn{1}{c}{\textbf{Den.}}\\  \hline
\multirow{2}{*}{DanceTrack \cite{sun2022dancetrack}}& Train &        1&        40&   41.8K       &    419      &   348.93K  & 8(5) & 89(9) & 20(17) & 76(8)  & 2.62(0.99)  \\ 
& Val   &        1&        25&    25.5K      &     273     &   225.15K     & 9(4) & 91(9) & 21(17) & 77(6)  & 2.74(0.98) \\ \midrule
\multirow{2}{*}{AnimalTrack \cite{zhang2022animaltrack}} & Train &        10 &        32 &          11.5K &    823 &         186K  & 16(9) & 91(14) & 15(15) & 71(8)  & 3.09(1.18) \\ 
            & Test   &        10&        26&          13.2K&          1.1K &         243K  & 19(7) & 92(8) & 15(15) & 72(7) & 3.17(1.26) \\ \midrule
\multirow{2}{*}{SportMOT \cite{cui2023sportsmot}}    & Train  &        1  &        45 &     28.57K     &        639  &    312.58K      & 11(3) & 80(16) & 18(16) & 73(8)  & 2.44(0.83) \\ 
& Val  &    1    &        45&      26.97K    &   641       &    295.57K      & 11(3) & 80(16) & 18(17) & 73(8) & 2.44(0.76) \\ \midrule
GMOT-40  \cite{bai2021gmot}      & Test  &        10 &        40&     9.64K     &    1.94K      &    256.34K     & 24(17)  & 43(44) & 11(12) & 71(9) & 2.56(0.88)  \\ \midrule
\textbf{G$^2$MOT(Ours)  }     & Test  &   20     &   253     &   157.2K       &    5.84K      &   1.87M       & 12(5)  & 84(14)  & 18(16) & 74(8)  & 2.65(0.95) \\ \bottomrule
\end{tabular}}
\label{tb:data_statistic}
\vspace{-10mm}
\end{table}


\section{G$^2$MOT Dataset}
\label{sec:dataset}
Ensuring a fair assessment of GMOT methods demands a dataset of consistent quality, free from annotator bias, and with a clearly defined problem setup. To offer comprehensive coverage of real-world scenarios across different research domains, our released dataset embodies two characteristics: (i) {\textit{Diversity:}} integrating diverse object categories from various sources, encompassing a broad spectrum of classes and diverse properties such as motion, occlusion, appearance similarity, and density. Additionally, it employs high-level semantics like \texttt{player}, \texttt{athlete}, \texttt{referee} etc., to describe objects in complex contexts, rather than using generic terms like \texttt{person}. (ii) {\textit{Fine-Grained Annotation: }} alongside capturing detailed visual attributes like color, texture, and attachments, it offers extensive textual descriptions with existing synonyms alongside captions.

\vspace{-3mm}
\subsection{Video collection}
\vspace{-2mm}
Combining datasets in object tracking offers strategic advantages. First, individual tracking datasets focus on specific challenges. Second, merging tracking datasets yields diverse challenges requiring tracking models to efficiently in varied scenarios. Therefore, by combining datasets, we can evaluate the tracking models' ability to deal with diverse scenarios e.g. object movements, density, similar appearance, and occlusion which are in line with the goal of the GMOT challenge. Finally, our ultimate objective is to \textit{propose a new paradigm} for GMOT and create a \textit{challenging benchmark dataset under various demanding real-world scenarios}.
Our G$^2$MOT dataset, a combination four benchmarks (Table \ref{tb:data_statistic}) not only broadens the range of object categories but also highlights the need for reliable tracking methods across different real-world contexts, e.g., \textit{generic categories diversity, fast-moving objects, high occlusion, long gaps, camera motion, etc}. 
\vspace{-3mm}
\subsection{Annotation}
\vspace{-2mm}
\label{sec:anno}

Our objective is to create a dataset with \textit{precise descriptions} and \textit{ambiguity-free annotations}, ensuring consistency for evaluation.
Our caption generation process is \textit{conducted manually}, emphasizing the need for careful attention to detail and accuracy. The annotation comprises two components: textual description annotation and tracking annotation. 

\begin{figure}
    \centering
    \includegraphics[width=\linewidth]{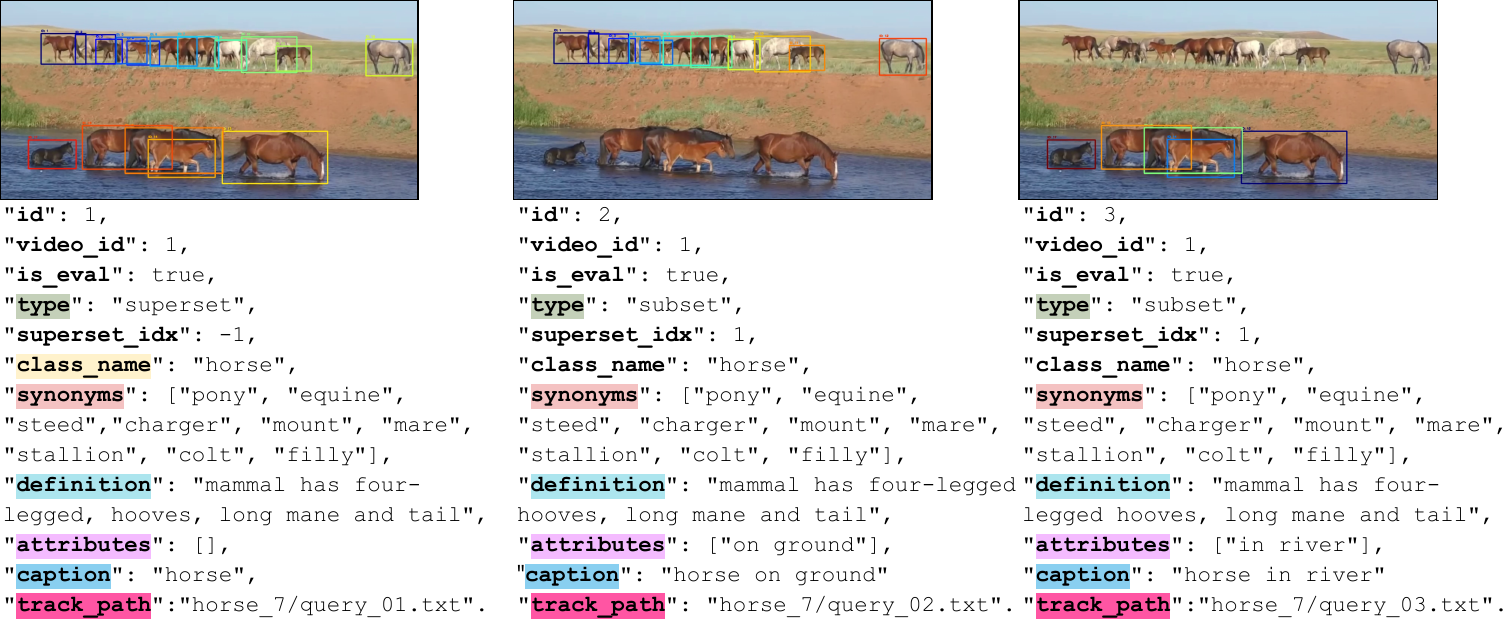}
    \caption{Demonstration of both \texttt{``superset''} and \texttt{``subset''} types within the same video and other fields in our annotation format which is described in Section \ref{sec:anno}.}
    \label{fig:fig_w_anno}
\end{figure}

\noindent
\underline{\textit{Textual description annotation:}} The textual description annotation, as shown in Figure \ref{fig:fig_w_anno}, is structured within JSON files and encompasses the following key fields:
\colorbox{classname}{\texttt{class\_name}}: represents the common name of the object class; 
\colorbox{type}{\texttt{type: superset | subset}}: indicates whether the object belongs to a ``superset" category, grouping ``coarse category" (e.g., horse), or a ``subset" category, allowing for finer categorization (e.g., horse on ground) as in Fig. \ref{fig:fig_w_anno}; 
\colorbox{caption}{\texttt{caption}}: manually crafted comprehensive description providing detailed information about the tracked objects;
\colorbox{synonym}{\texttt{synonyms}}: Offers alternative terms or phrases for the class name; 
\colorbox{definition}{\texttt{definition}}: describe the object's visual characteristics.
\colorbox{attribute}{\texttt{attributes}}: encompasses a list of attributes used to distinguish objects within the ``superset'';
\colorbox{track_path}{\texttt{track\_path}}: follows the standard MOT format and is stored separately.

\noindent
\underline{\textit{Tracking annotation:}} The tracking annotation follows the standard MOT format \cite{milan2016mot16,dendorfer2020mot20} includes the following parameters [\texttt{frame\_id,} \texttt{identity\_id,} \texttt{box\_top\_left\_x,} \texttt{box\_top\_left\_y,} \texttt{box\_width,} \texttt{box\_height,} \texttt{1, -1, -1, -1}]. 

Equipped with comprehensive annotations and diverse attributes, our G$^2$MOT dataset extends its utility beyond object tracking to support various grounding tasks e.g., Question Answering, fine-grained task understanding in real-world scenarios. Moreover, G$^2$MOT provides adaptability in tracking objects under different configurations, including captions (default), attributes, object definitions, and synonyms, effectively addressing the complexities of real-world natural language descriptions. Further detailed data description is included in the Supplementary Material (Supp.).

\begin{figure}
    \centering
    \begin{subfigure}[t]{0.3\linewidth}
        \includegraphics[width=0.9\linewidth]{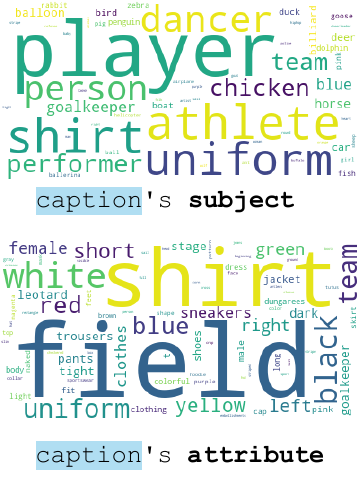}
        \caption{Demonstration of word use in the \colorbox{babyblue}
     {\texttt{caption}} under Word cloud in G$^2$MOT dataset.}
    \end{subfigure}
    \hfill
    \begin{subfigure}[t]{0.6\linewidth}
        \includegraphics[width=\linewidth]{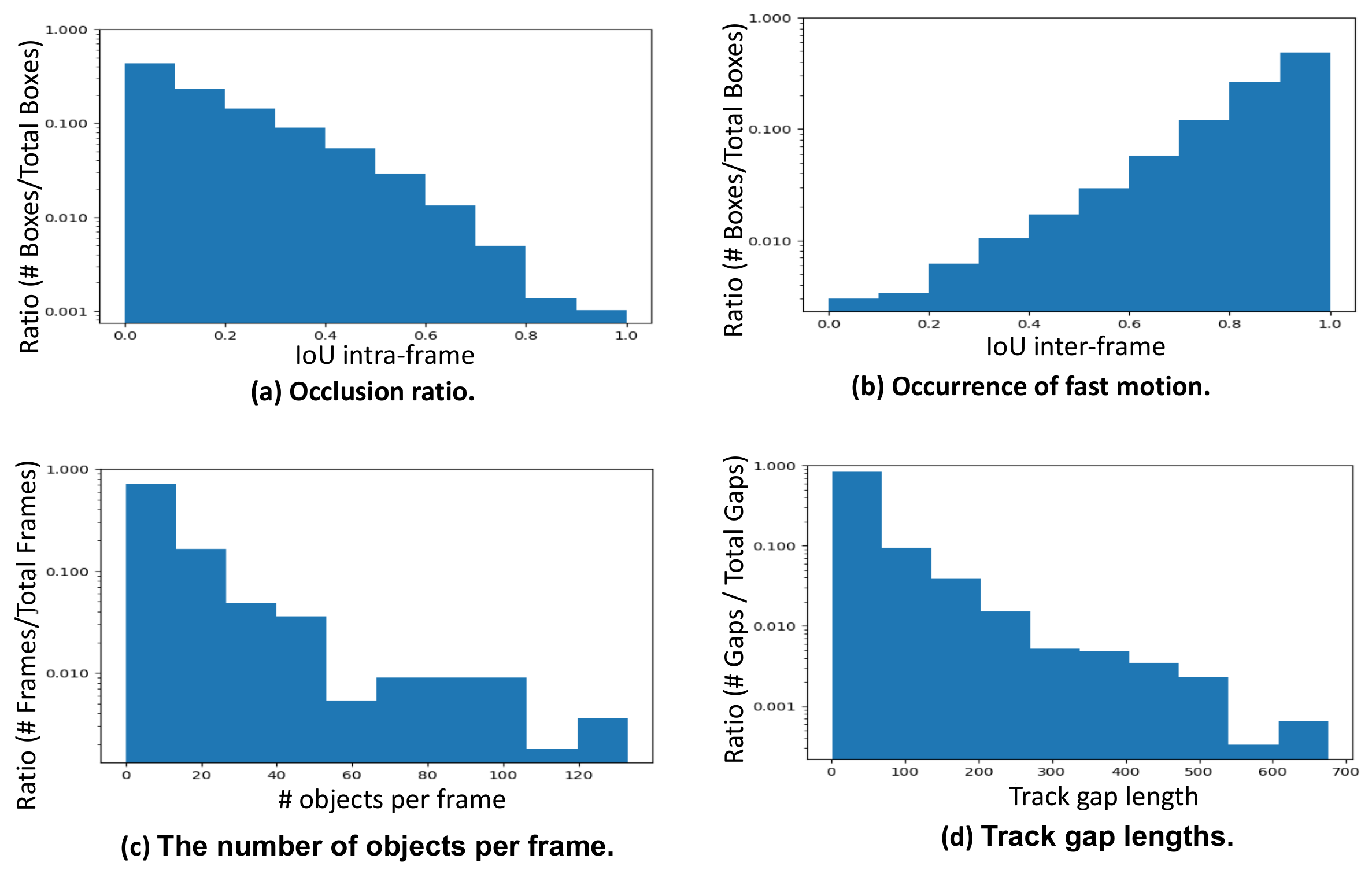}
        \caption{Demonstration of a wide range of statistical information of G$^2$MOT, highlighting characteristics such as occlusions (a), fast-moving objects (b), a high number of tracklets (c), and tracking gaps (d).}
    \end{subfigure}
\caption{Statistical information of our proposed G$^2$MOT.}
\label{fig:data_statistic}
\end{figure}

\subsection{Data statistics}
Table \ref{tb:dataset_comparison} provides a comprehensive comparison with existing datasets, while Table \ref{tb:data_statistic} offers detailed information about our G$^2$MOT. Among GMOT datasets, ours has the highest number of categories and videos, surpassing all MOT datasets except for TAO. However, it's important to note that TAO, despite having a higher video count, lacks dense annotation and exhibits lower annotation quality, with fewer challenges such as low appearance similarity, and less occlusion.
Additionally, while some datasets such as MOT20 \cite{dendorfer2020mot20} contain a high number of objects per frame (Obj.) but low appearance similarity (App.), less occlusion (Occ.), slow motion (Mot.), and DanceTrack \cite{sun2022dancetrack} exhibit high App., substantial Occ., but slower motion (Mot.), fewer Obj., our G$^2$MOT, being a combination of multiple datasets, provides a diversity of challenges including a large Obj., high App., dense object density (Den.), substantial Occ., and fast Mot. 
While existing referring datasets \cite{fan2019lasot,wang2021towards,wu2023referring} only provide captions as tracking settings and present a low range of data properties, including low scores of Obj., App., Den., Occ., and Mot., our G$^2$MOT offers fine-grained information with various textual description settings including definition, attributes, synonyms, besides captions, and contains a wide range of diversity in data properties. This is depicted in Tables \ref{tb:dataset_comparison} and \ref{tb:data_statistic} through the dataset's statistical information and data properties, including mean and standard deviation on each metric. Detailed computation of these metrics (Obj., App., Den., Occ., Mot.) is included in the Supp.

Figure \ref{fig:data_statistic}(a) presents a Word Cloud that depicts the frequent usage of terms related to the \colorbox{babyblue}{\texttt{caption}}'s \textit{subject} and \colorbox{babyblue}{\texttt{caption}}'s \textit{attributes}. Not only diversity within the object category, but it also diversifies in higher semantic levels e.g., ``player'', ``athlete'', ``dancer'' rather than ``person'' only. Regarding the attribute part, the most frequently occurring descriptors encompass color, object parts, and locations. Figure \ref{fig:data_statistic}(b) summarizes some further attributes, including (a) occlusion: measured by the IoU of interacting objects; (b) occurrence of fast motion: determined by the IoU between object boxes in two adjacent frames; (c) number of objects per frame: calculated by number of object in a frame. (d) track gap lengths: measured by the gap length between the frames in which an object reappears and the frame of its last occurrence.

\vspace{-3mm}
\subsection{Grounded-GMOT benchmark protocols}
\vspace{-2mm}
In this evaluation process, we make use of well-established metrics as defined in \cite{wu2023referring,sun2022dancetrack}. Specifically, we employ the following metrics: Higher Order Tracking Accuracy ($HOTA$) \cite{Luiten_2020hota}, Multiple Object Tracking Accuracy ($MOTA$) \cite{Bernardin2008mota}, and $IDF1$ \cite{Ristani2016idf1}, together with Detection Accuracy ($DetA$), Association Accuracy ($AssA$). It is important to note that $HOTA = \sqrt{DetA \cdot AssA}$ effectively strikes a balance in assessing both frame-level detection and temporal association performance.

\section{Proposed KAM-SORT}

\SetKwInput{KwModel}{Model}
\SetKwInput{KwOutput}{Output}

\begin{wrapfigure}{r}{0.55\textwidth}
\vspace{-4mm}
\scriptsize 
\begin{algorithm}[H]
\caption{\small Kalman++ algorithm} 
\vspace{-5mm}
\label{alg:cap}
\KwData{ \\
    $\mathcal{D}$, $\mathcal{T}$: set of detection boxes at current frame and tracks at the previous frame. \\
     $\alpha$: param. of uncertainty revise factor.
}
\KwModel{ \\
    $C$: score matrix defined in Equation \ref{eq:cost_1}; 
    $M$: bipartite matching function; 
    $K_p$, $K_u$: Kalman Filter predict and update; 
    $BC$, $IoU$: function compute box center and IoU.
}
\KwOutput{ \\
    $\mathcal{T'}$ set of new tracks. \\
}

$\hat{x}, P = K_p(\mathcal{T});$ \tcp{Get estimated location and error covariance.}

$S$ = $C(\hat{x}, \mathcal{D});$ \tcp{Compute matching score between estimation and detection.} 

$DT_m, D_r, T_r$ = $M(S)$; \tcp{$1^{st}$-round association produce matched pairs $DT_m$, unmatched detections $D_r$, and unmatched tracks $T_r$.}

$S_{IoU}$ = $IoU(D_r, T_r);$ \tcp{$2^{nd}$-round associate unmatched ones.}

$DT_r$ = $M(S_{IoU});$ \tcp{Rematched pairs from remaining detections and tracks.}

\For{$(i_d, i_t) \in DT_r$}{ \tcp{$i_d$: detection index, $i_t$: track index.}
    $c^{min}$ = $\hat{x}_{i_t}[:2] - \alpha \sqrt{P[:2]}$ \text{ and }
    $c^{max}$ = $\hat{x}_{i_t}[:2] + \alpha \sqrt{P[:2]}$; \text{ and }
    $c$ =  $BC(\mathcal{D}_{i_d})$; \\
    \If{$c > c^{min} \& c < c^{max}$}{
     $DT_m$ = $DT_m \cup (i_d, i_t);$ 
    }
}  

$\mathcal{T'}$ = $K_u(DT_m)$ \tcp{Update matched tracks}
\end{algorithm}
\end{wrapfigure}

MOT is primarily designed to track objects with diverse appearances, such as individuals wearing various outfits or having distinct facial features and hairstyles. In contrast, GMOT is tailored for tracking objects of a generic type that share a high degree of visual similarity. This task becomes particularly challenging when attempting to associate objects across consecutive frames, especially in scenarios where objects densely congregate, as seen in groups like schools of fish, ant colonies, or swarms of bees. Consequently, our approach advocates for the utilization of both visual representations and motion cues to effectively track these generic objects. Our proposed KAM-SORT consists of two major contributions: (i) propose a tracking mechanism that can dynamically balance visual appearance and motion cues (ii) propose \textbf{Kalman++}, an improvement to the Kalman filter to re-associate unmatched detections and unmatched tracks.

\label{sec:kamsort}

The problem's setting is as follows: Consider a set of $N$ existing tracks $\mathcal{T}$ and a set of $M$ new detections in the current time step $\mathcal{D}$. The standard similarity between the track $\mathcal{T}$ and box embeddings $\mathcal{D}$ is defined by cosine distance and is represented as $C_a \in \mathbb{R}^{M\times N}$. In a typical tracking approach that combines visual appearance and motion cues, the cost matrix $C$ is:
\begin{equation}
   C(\mathcal{T}, \mathcal{D}) = C_m(\mathcal{T}, \mathcal{D}) + \gamma C_a(\mathcal{T}, \mathcal{D}).
\end{equation}
Here, $C_m$ represents the motion cost, which is measured using the IoU cost matrix. Leveraging OC-SORT, a technique that calculates a virtual trajectory over occlusion periods to correct error accumulation in filter parameters during occlusions, the motion cost is defined as:
\begin{equation}
C_m(\mathcal{T}, \mathcal{D}) = IoU(\mathcal{T}, \mathcal{D}) + \lambda C_v(\tilde{T}, \mathcal{T}).
\end{equation}
Therefore, the resulting cost matrix that integrates both visual appearance and motion information is as follows:
\begin{equation}
C(\mathcal{T}, \mathcal{D}) = IoU(\mathcal{T}, \mathcal{D}) + \lambda C_v(\tilde{T}, \mathcal{D}) + \gamma C_a(\mathcal{T}, \mathcal{D}).
\label{eq:cost_0}
\end{equation}
, where $\tilde{T}$ contains the trajectory of observations of all existing tracks. $C_v$ represents the consistency between the directions of i) linking two observations (i.e., $\mathcal{T}^{t-2}$, $\mathcal{T}^{t-1}$: a set of tracks at time $(t-2)$  $(t-1)$), denote as, $(\mathcal{T}^{t-2}\rightarrow\mathcal{T}^{t-1})_{[u,v]}$ and ii) linking tracks' historical observations $\mathcal{T}^{t-1}$ and new observations $\mathcal{D}$ at frame $t$, $(\mathcal{T}^{t-1}\rightarrow \mathcal{D})_{[u,v]}$. As a result, $C_v$ is computed on 2D coordinates $[u, v]$ of the object center:
\begin{equation}
    C_v = \mathtt{arctan}\left((\mathcal{T}^{t-2}\rightarrow \mathcal{T}^{t-1})_{[u,v]}, (\mathcal{T}^{t-1}\rightarrow \mathcal{D})_{[u,v]}\right).
\end{equation}

To strike a balance between visual appearance and motion cues, we incorporate adaptive appearance cost $W_a$ and adaptive motion cost $W_m$ into Equation \ref{eq:cost_0}, resulting in:
\begin{equation}
C(\mathcal{T}, \mathcal{D}) = W_mIoU(\mathcal{T}, \mathcal{D}) + \lambda C_v(\tilde{T}, \mathcal{D}) + W_aC_a(\mathcal{T}, \mathcal{D}).
\label{eq:cost_1}
\end{equation}
To effectively handle the high similarity between objects of the same generic type in GMOT, we propose the following hypothesis: when the visual appearances of all detections are very similar, the tracker should prioritize motion over appearance. The homogeneity of visual appearances across all detections can be quantified as follows:
\begin{equation}
\mu = \frac{1}{M}\sum_{i=1}^{M}{f_i} \text{ and }
\mu_{det} = \frac{1}{M}\sum_{i=1}^{M}{\mathtt{cos}(f_i, \mu)}.
\label{eq:distribution}
\end{equation}
Here, we consider a threshold $\theta$ to determine the similarity between two vectors; if the angle between them is smaller than $\theta$, the vectors are considered more similar. It's noteworthy that when $\mu_{det} > \mathtt{cos}(\theta)$, the visual appearance is less reliable for tracking, implying that $C_{a}$ should be less than 1. Conversely, $C_{a} > 1$ when $\mu_{det} < \mathtt{cos}(\theta)$. Therefore, the weight $C_{a}$ can be calculated as:
    \begin{equation}
    W_a = \frac{(1 -\mu_{det})}{1-\mathtt{cos}(\theta)}.
\label{eq:Wa}
\end{equation}
We initialize $C_{m}$ as 1, indicating that both motion and appearance are equally important. As the weight on appearance reduces, we propose redistributing the remaining weight to motion. Thus, the adaptive motion weight $C_{m}$ is:
\begin{equation}
    W_m = 1 + \left[1 - W_a\right] = 2 - \frac{(1 -\mu_{det})}{1-\mathtt{cos}(\theta)}.
\end{equation}
As a result, the final cost matrix $C$ is computed as follows:
\begin{equation}
\small
C(\mathcal{T}, \mathcal{D}) = \left(2 - \frac{(1 -\mu_{det})}{1-\mathtt{cos}(\theta)}\right)IoU(\mathcal{T}, \mathcal{D})
+ \lambda C_v(\tilde{T}, \mathcal{D}) 
+  \frac{(1 -\mu_{det})}{1 -\mathtt{cos}(\theta)}C_a(\mathcal{T}, \mathcal{D}).
\label{eq:cost_2}
\end{equation}

In our KAM-SORT framework, the cost matrix between existing tracks $\mathcal{T}$ and detections $\mathcal{D}$ is computed using our proposed Kalman++ algorithm, outlined in Algorithm \ref{alg:cap}. Specifically, we introduce an uncertainty revision parameter ($\alpha$) to re-associate unmatched detections and tracks. From our observations, we have noticed a significant variation in box centers when dealing with fast motion and object deformation. This variation introduces unwanted noise in linear estimators like the standard Kalman Filter, leading to mismatches between detections and tracks. As a result, we propose to employ IoU scores to associate detections with previously unmatched tracks. Our Kalman++ algorithm strategically adjusts the probabilistic bounding box by considering the variance in predictions. This adjustment allows for the expansion or contraction of the predicted bounding box based on the variance of the prediction, providing flexible thresholds ($c_{min}, c_{max}$) that adapt to the level of uncertainty in the prediction.

\vspace{-5mm}
\section{Experimental Results}
\vspace{-2mm}
\subsection{Implementation Details}
\vspace{-2mm}

\vspace{-5mm}
\begin{figure*}[]
    \centering
    \includegraphics[width=\linewidth]{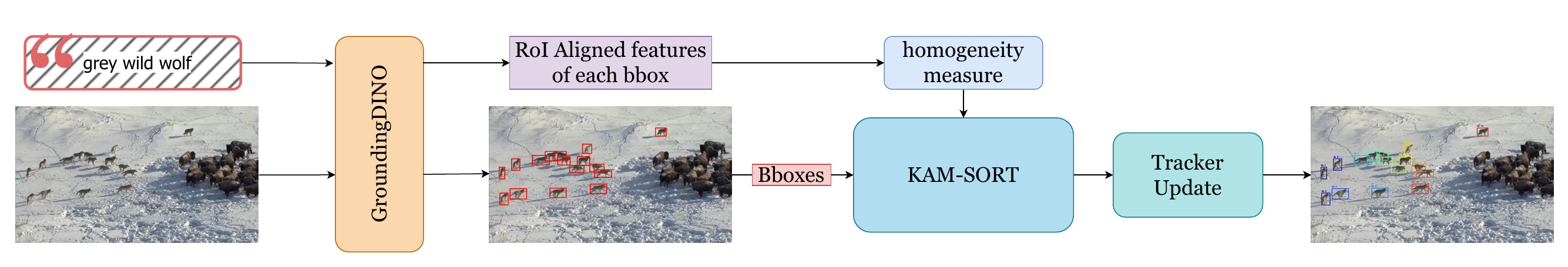}
    \vspace{-5mm}
    \caption{\small Overview of the proposed Grounded-GMOT pipeline. The system detects objects using Grounding DINO with natural language descriptors (e.g., "grey wild wolf"), measure homogeneity via RoI-aligned features, and uses KAM-SORT for object association, enabling zero-shot multi-object tracking without prior training.
}
    \label{fig:pipeline}
\end{figure*}
\vspace{-5mm}

In the context of GMOT, the state-of-the-art (SOTA) approach is known as OS-GMOT \cite{bai2021gmot}. To ensure fair comparisons, Grounded-GMOT is evaluated against OS-GMOT. We implement OS-GMOT following the configuration outlined in SOTA \cite{bai2021gmot}. For Grounded-GMOT, we utilize GroundingDINO \cite{liu2023grounding} alongside \colorbox{caption}{\texttt{captions}} from the proposed G$^2$MOT dataset to generate detected bounding boxes. As shown in Fig.~\ref{fig:pipeline}, the system detects objects using natural language descriptors (e.g., "grey wild wolf"), calculate homogeneity across objects via RoI-aligned features, and passes them to KAM-SORT for object association across frames. OS-GMOT operates on a one-shot basis, whereas our Grounded-GMOT employs a zero-shot tracking mechanism, with no training required in either OS-GMOT or Grounded-GMOT.


To evaluate the efficacy of our KAM-SORT, we compared it with several established trackers, including SORT \cite{bewley2016simple}, DeepSORT \cite{wojke2017simple}, BYTETrack \cite{zhang2021bytetrack}, OC-SORT \cite{cao2023observation}, Deep OCSORT \cite{maggiolino2023deep}, MOTRv2 \cite{zhang2023motrv2}. It is worth noting that while KAM-SORT is SORT-based, MOTRv2 utilizes a transformer-based architecture. 
To implement KAM-SORT, we configured the parameter with uncertainty revision $\alpha = 1$ and similarity threshold $\theta = 80^\circ$  in our experiments.




\vspace{-3mm}
\subsection{Performance Comparison}

\begin{table*}[]
  \centering
  \vspace{-3.1em}
  \setlength{\tabcolsep}{5pt}
    \renewcommand{\arraystretch}{1}
      \caption{\textit{Tracking performance} comparison of multiple trackers under various settings of MOT with YOLOv8 \cite{yolo16}, OS-GMOT (averaged over five runs), and our proposed \textit{Grounded-GMOT} on the G$^2$MOT dataset. The best score is in \textbf{bold}}
    \resizebox{\linewidth}{!}{
        \begin{tabular}{l|c|c|c|ccccc}
        \toprule
        \textbf{Trackers} & \multicolumn{2}{c|}{\textbf{Settings}} & \textbf{HOTA}$\uparrow$   & \textbf{MOTA}$\uparrow$          & \textbf{IDF1}$\uparrow$   & \textbf{DetA}$\uparrow$  & \textbf{AssA}$\uparrow$\\ \midrule
        \multirow{4}{*}{SORT \cite{bewley2016simple} } 
        & YOLOv8 & Fully-train & 5.48 & -145.61 & 0.80 & 5.78 & 6.47 \\
        & OS & Five runs of OS & 24.77 & 7.09 & 24.90 & 30.22 & 20.70 \\
        &\multirow{1}{*}{\textbf{Grounded-GMOT}}& \textbf{Zero-shot}  & \textbf{40.73} & \textbf{46.57} & \textbf{44.52} & \textbf{45.13} & \textbf{37.26} \\ \midrule
        \multirow{4}{*}{DeepSORT \cite{wojke2017simple} }
        & YOLOv8 & Fully-train& 5.21 & -156.2 & 0.74 & 5.88 & 5.82 \\
        & OS & Five runs of OS& 22.59 & -0.20 & 21.66 & 29.3 & 17.89 \\
        & \textbf{Grounded-GMOT} & Zero-shot & \textbf{36.01} & \textbf{43.30} & \textbf{37.54} & \textbf{43.94} & \textbf{29.96} \\ \midrule
        \multirow{4}{*}{ByteTrack \cite{zhang2021bytetrack} } 
        &  YOLOv8 & Fully-train &6.02 & -140.81 & 0.84 & 5.80 & 7.53 \\
        &OS-GMOT & Five runs of OS& 25.16 & 8.02 & 26.46 & 29.38 & 21.94 \\
        &\textbf{ Grounded-GMOT} & \textbf{Zero-shot} & \textbf{39.89} & \textbf{45.83 }& \textbf{45.65} & \textbf{43.35} & \textbf{37.12} \\ \midrule
        \multirow{4}{*}{OC-SORT  \cite{cao2023observation} } 
        & YOLOv8 &  Fully-train&  5.48 & -127.3 & 0.76 & 5.53 &  6.78 \\
        &OS-GMOT & Five runs of OS& 25.17 & 12.62 & 25.96 & 29.66 & 21.67 \\
        & \textbf{Grounded-GMOT} & \textbf{Zero-shot} & \textbf{41.84} & \textbf{46.32} & \textbf{45.92} & \textbf{44.49} & \textbf{39.92} \\ \midrule
        \multirow{4}{*}{Deep OCSORT  \cite{maggiolino2023deep} } 
        & YOLOv8 & Fully-train& 5.72 & -145.6 & 0.81 & 5.80 & 6.94 \\
        &OS-GMOT & Five runs of OS& 25.65 & 7.06 & 25.92 & 30.47 & 21.92 \\
        & \textbf{Grounded-GMOT}&\textbf{ Zero-shot }& \textbf{40.53} & \textbf{46.12}  & \textbf{43.08} &\textbf{ 46.01} & \textbf{36.27} \\\midrule
        \multirow{4}{*}{MOTRv2  \cite{zhang2023motrv2} } 
        & YOLOv8 & Fully-train& 3.06 & 0.48 & 0.85 & 0.45 & 20.71 \\
        &OS-GMOT & Five runs of OS& 28.69	&14.18	&29.43	&26.32	& 34.88 \\
        & \textbf{Grounded-GMOT}& \textbf{Zero-shot} & \textbf{42.02}	& \textbf{41.68}	& \textbf{45.91}	& \textbf{41.81}	& \textbf{42.54} \\ \bottomrule
        \end{tabular}
    }
    \label{tb:grounded-gmot}
\end{table*}
\vspace{-1em}


\begin{table}[!t]
  \centering
  \vspace{-1em}
    \setlength{\tabcolsep}{7pt}
    \renewcommand{\arraystretch}{1}
  \caption{\textit{Tracking performance} comparison between the \textit{existing trackers} and our proposed \textit{KAM-SORT} tracker on G$^2$MOT dataset. The best score is in \textbf{bold}.}
\resizebox{\linewidth}{!}{
\begin{tabular}{l|c|ccccc}
\toprule
\textbf{Trackers} & \textbf{Settings}& \textbf{HOTA}$\uparrow$   & \textbf{MOTA}$\uparrow$          & \textbf{IDF1}$\uparrow$   & \textbf{DetA}$\uparrow$  & \textbf{AssA}$\uparrow$\\ \midrule
SORT \cite{bewley2016simple} & Grounded-GMOT& 40.73 & 46.57 & 44.52 & 45.13 & 37.26 \\
DeepSORT \cite{wojke2017simple} & Grounded-GMOT& 36.01 & 43.30 & 37.54 & 43.94 & 29.96 \\
ByteTrack \cite{zhang2021bytetrack} & Grounded-GMOT& 39.89 & 45.83 & 45.65 & 43.35 & 37.12 \\
OC-SORT \cite{cao2023observation}  & Grounded-GMOT& 41.84 & 46.32 & 45.92 & 44.49 & 39.92 \\ 
DeepOC-SORT\cite{maggiolino2023deep} & Grounded-GMOT & 40.53 & 46.12  & 43.08 & 46.01 & 36.27 \\
MOTRv2 \cite{zhang2023motrv2} & Partly-trained &42.02	&41.68	&45.91	&41.81	&\textbf{42.54}\\ \midrule
\textbf{KAM-SORT(Ours)} & Grounded-GMOT  &\textbf{43.03} & \textbf{46.60} & \textbf{47.13} & \textbf{46.05} & 40.80 \\ \bottomrule

\end{tabular}}
\vspace{-1em}
\label{tb:assoc_phase}
\end{table}

\noindent
\underline{\textit{Comparing Grounded-GMOT with OS-GMOT and full-trained trackers. }}
We evaluate the efficacy of the Grounded-GMOT paradigm by comparing it with (i) OS-GMOT averaged over five runs and (ii) traditional MOT settings where the object detector is fully-trained, specifically YOLOv8 \cite{yolo16} trained on MSCOCO \cite{Lin2014mscoco}, using various trackers as detailed in Table \ref{tb:grounded-gmot}. This experiment highlights Grounded-GMOT's advancements in GMOT. 

\begin{wraptable}{r}{0.5\textwidth}
\centering
\vspace{-3mm} 
\setlength{\tabcolsep}{2pt}
\renewcommand{\arraystretch}{0.9}
  \caption{\small \textit{Tracking performance} of \textit{KAM-SORT} on G$^2$MOT with \textit{various settings}.}
  \vspace{-2mm}
\resizebox{\linewidth}{!}{
\begin{tabular}{l|ccccc}
\toprule
\multicolumn{1}{c}{\textbf{Settings}}& \textbf{HOTA}$\uparrow$   & \textbf{MOTA}$\uparrow$          & \textbf{IDF1}$\uparrow$   & \textbf{DetA}$\uparrow$  & \textbf{AssA}$\uparrow$\\ \midrule
\colorbox{attribute}{\texttt{attribute}} + \colorbox{classname}{\texttt{class\_name}} & 42.20 & 43.26 & 45.29 & 44.73 & 40.15 \\ 
 \colorbox{definition}{\texttt{definition}}& 34.04 & 26.45 & 35.83 & 34.00 & 34.49 \\
 \colorbox{babyblue}{\texttt{caption}}& \textbf{43.03} & \textbf{46.60} & \textbf{47.13} & \textbf{46.05} & \textbf{40.80} \\ \bottomrule
\end{tabular}}
\label{tb:various_settings}
\end{wraptable}

\begin{wraptable}{r}{0.4\linewidth}
\centering
\vspace{-31mm}
\setlength{\tabcolsep}{2pt}
\renewcommand{\arraystretch}{0.9}
  \caption{\scriptsize Ablation study on the \textit{effectiveness} of \textit{KAM-SORT} on \textit{MOT20-testset} with MOT task. As \textcolor{gray}{ByteTrack, OC-SORT} uses different thresholds for testset sequences with an offline interpolation procedure, we also report scores by disabling these as in ByteTrack$^\dag$, OC-SORT$^\dag$. As \textcolor{gray}{Deep OC-SORT} used separated weights for YOLOX, we also report scores by retraining YOLOX on MOT20-trainset as in Deep OC-SORT$^\dag$.}

\resizebox{\linewidth}{!}{
\begin{tabular}{l|ccc}
\toprule
\textbf{Trackers}  &  \textbf{HOTA}$\uparrow$ & \textbf{MOTA}$\uparrow$ & \textbf{IDF1}$\uparrow$  \\ \midrule
MeMOT\cite{cai2022memot}  & 54.1 & 63.7 & 66.1     \\ 
 FairMOT\cite{zhang2021fairmot} & 54.6 & 61.8 & 67.3     \\   
 GSDT\cite{Wang2020_GSDT}   & 53.6 & 67.1 & 67.5   \\  
  CSTrack\cite{liang2022cstrack}     & 54.0 & 66.6 &  68.6 \\  
 \color{gray}{ByteTrack}\cite{zhang2021bytetrack}      & \color{gray}61.3 & \color{gray}{77.8}  & \color{gray}75.2   \\ 
 \color{gray}OC-SORT\cite{cao2023observation}    &  \color{gray}62.4	&\color{gray}75.7 &	\color{gray}76.3   \\ 
  \color{gray}{Deep-OCSORT}\cite{maggiolino2023deep}  & \color{gray}63.9 &\color{gray}75.6   &\color{gray}79.2    \\ 
 ByteTrack$^\dag$\cite{zhang2021bytetrack}     & 60.4 & 74.2  & 74.5 \\ 
 OC-SORT$^\dag$\cite{cao2023observation}   & \underline{60.5} & 73.1  & 74.4  \\ 
  Deep OC-SORT$^\dag$\cite{maggiolino2023deep} &59.6 & \textbf{75.3} &\underline{75.2} \\ \midrule
 {\textbf{KAM-SORT} \textbf{(Ours)}} & \textbf{62.6} & \underline{75.2}  & \textbf{76.9} \\
\bottomrule
\end{tabular}
}
\label{tb:Abl_MOT20}
\end{wraptable}

Particularly, Grounded-GMOT demonstrates superior object detection capabilities (higher MOTA and DetA scores) attributed to the robust grounding capabilities of VLM. Additionally, leveraging textual descriptions from \colorbox{caption}{\texttt{captions}}, Grounded-GMOT reduces reliance on initial bounding boxes, a challenge in OS-GMOT. Furthermore, while YOLOv8 \cite{yolo16} trained on MSCOCO \cite{Lin2014mscoco} fails to detect categories not present in the training set, Grounded-GMOT outperforms all OS-GMOT and fully-trained MOT approaches in handling the GMOT task.

\noindent
\underline{\textit{Compare KAM-SORT with SOTA MOT methods. }}

\begin{figure*}[]
    \centering
    \includegraphics[width=\linewidth]{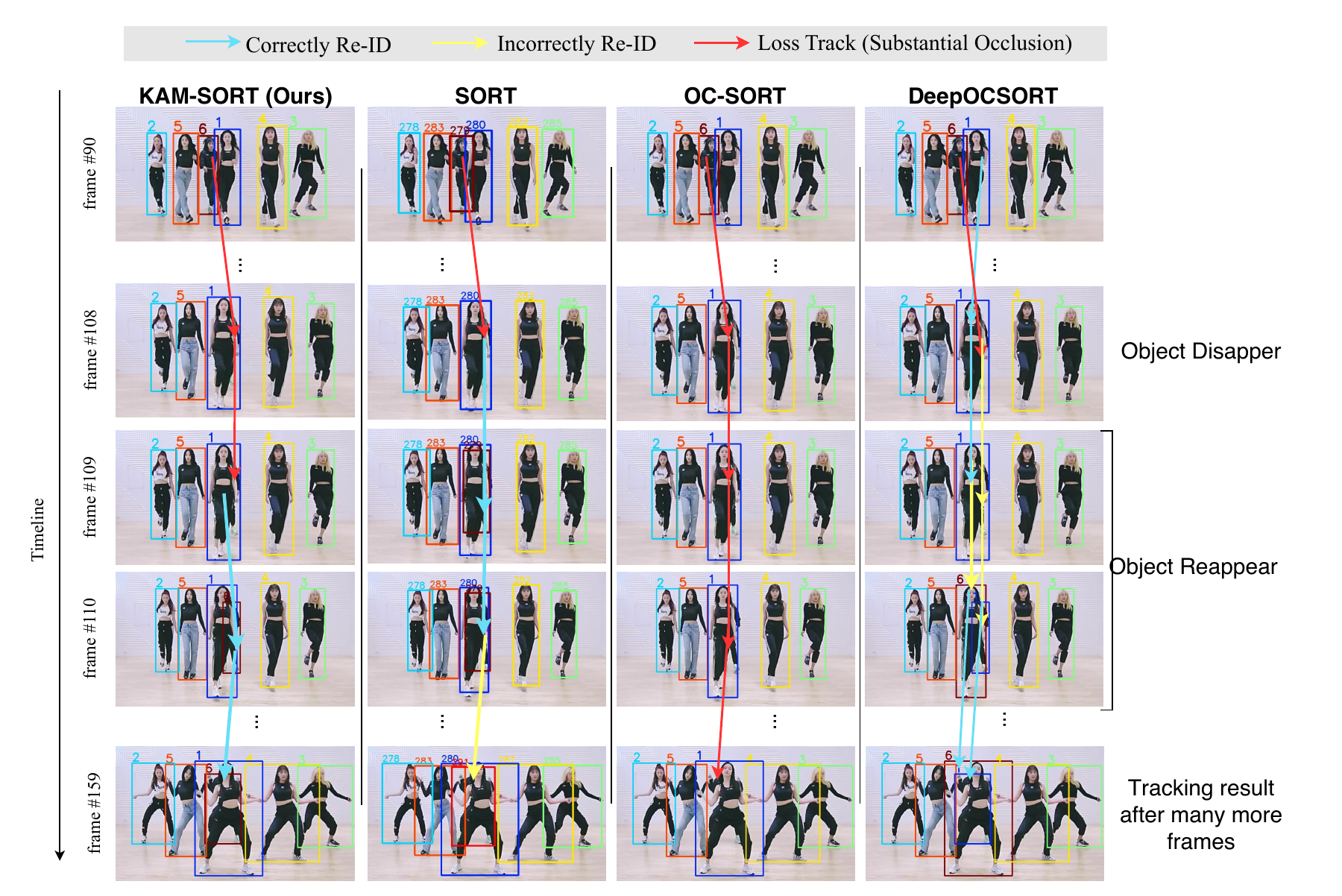}
    \vspace{-5mm}
    \caption{\small Tracking comparison between our tracker KAM-SORT ($1^{st}$ column) with SORT ($2^{nd}$ column), OC-SORT ($3^{rd}$ column) and DeepOCSORT ($4^{th}$ column) on video dancetrack0010 \cite{sun2022dancetrack}. When handling objects disappear and reappear, SORT encounters challenges in maintaining tracklets, OC-SORT tends to lose track, potentially leading to incorrect Re-ID, and DeepOCSORT faces difficulties Re-ID objects once they reappear. In contrast, our KAM-SORT accurently re-associates object ID once the object reappears.}
    \label{fig:fig_trackers_compare}
\end{figure*}

Table \ref{tb:assoc_phase} provides a thorough comparison between our proposed KAM-SORT and SOTA existing trackers. Our method is primarily SORT-based, necessitating an evaluation against other SORT-based approaches. In this experiment, SORT-based trackers adhere to the Grounded-GMOT setting, utilizing object detections by GroundingDINO-B. However, for a comprehensive perspective, we have also included MOTRv2 \cite{zhang2023motrv2}, a cutting-edge transformer-based tracker, as a reference point in this assessment. The results demonstrate that KAM-SORT outperforms all trackers including both SORT-based and transformer-based ones across metrics except MOTRv2 on AssA score. It is noteworthy that MOTRv2 was partially trained on the G$^2$MOT dataset, specifically on the training set of DanceTrack dataset\cite{sun2022dancetrack}. Additionally, we conduct a visual comparison between our KAM-SORT and other trackers, as depicted in Fig. \ref{fig:fig_trackers_compare}. In this illustration, SORT encounters challenges with loss track and incorrectly re-ID, resulting in numerous new IDs being associated at Frame \#90. Let's consider the object with ``ID = 6'' in other trackers, OC-SORT struggles with ID Re-association caused by loss track, while Deep-SORT faces issues with ID switching and incorrectly Re-ID. In contrast, our KAM-SORT accurately re-associates object ID once the object reappears.

\vspace{-5mm}
\begin{table}[h]
    \centering
    \begin{minipage}[t]{0.53\textwidth}
        \centering
        \caption{An ablation study conducted on the G$^2$MOT dataset to demonstrate the \textit{impact of each proposed component} within \textit{KAM-SORT}.}
        \vspace{-3pt}
        \resizebox{\textwidth}{!}{%
        \begin{tabular}{l|c|c|ccccc}
        \toprule
        \multirow{2}{*}{\textbf{Exp.}} & {\textbf{Appearance-}} & \multirow{2}{*}{\textbf{Kaman++}} & \multicolumn{5}{c}{\textbf{Tracking Metrics}} \\
        & \textbf{Motion Balance}  & & {HOTA}$\uparrow$   & {MOTA}$\uparrow$          & {IDF1}$\uparrow$   &{DetA}$\uparrow$  & {AssA}$\uparrow$\\
        \midrule
         \texttt{\#1} & \xmark & \xmark & 40.53 & 46.12 &43.08 & 46.01 & 37.27 \\ 
         \texttt{\#2} & \xmark & \cmark & 41.90 & 46.35 & 45.27 & 46.02 & 38.71\\
         \texttt{\#3} & \cmark & \xmark & 43.03 & 46.60 & 47.12 & 46.05 & 40.79\\
         \texttt{\#4} & \cmark & \cmark & {43.03} & 46.60 & 47.13 & 46.05 & 40.80\\
        \bottomrule
        \end{tabular}}
        \label{tab:abl-kamsort}
    \end{minipage}%
    \hfill
    \begin{minipage}[t]{0.45\textwidth}
        \centering
        \caption{Ablation study on \textit{hyper-parameters} on \textit{KAM-SORT}.}
        \vspace{9pt}
        \resizebox{\textwidth}{!}{%
        \begin{tabular}{l|ccc||l|ccc}
        \toprule
        \multicolumn{4}{c||}{\textbf{Vector Similarity }$\theta$} & \multicolumn{4}{c}{\textbf{Uncertainty Revision} $\alpha$} \\ \hline
        $\theta$ & HOTA$\uparrow$   & MOTA$\uparrow$  & IDF1$\uparrow$    & $\alpha$ & HOTA$\uparrow$      &  MOTA$\uparrow$ & IDF1$\uparrow$   \\ \hline
        22.5$^\circ$ & 42.963 & 46.586 & 47.013 &  0.5 & 43.026 &46.601 & 47.123 \\
        45$^\circ$ & 43.010 & 46.600 & 47.091 & 1 & 43.027 &46.601 & 47.126  \\   
        67.5$^\circ$ & 43.020 &46.601 & 47.122 & 2 & 43.026 &46.602 & 47.131  \\
        80$^\circ$ &  43.027 & 46.601 & 47.126 & 3 & 43.026 & 46.602 & 47.131 \\
        \bottomrule
        \end{tabular}}
        \label{tab:ablation_all}
    \end{minipage}
\end{table}
\vspace{-5mm}

\vspace{-3mm}
\subsection{Ablation Study}
\vspace{-2mm}

\begin{figure*}
    \centering
    \includegraphics[width=\linewidth]{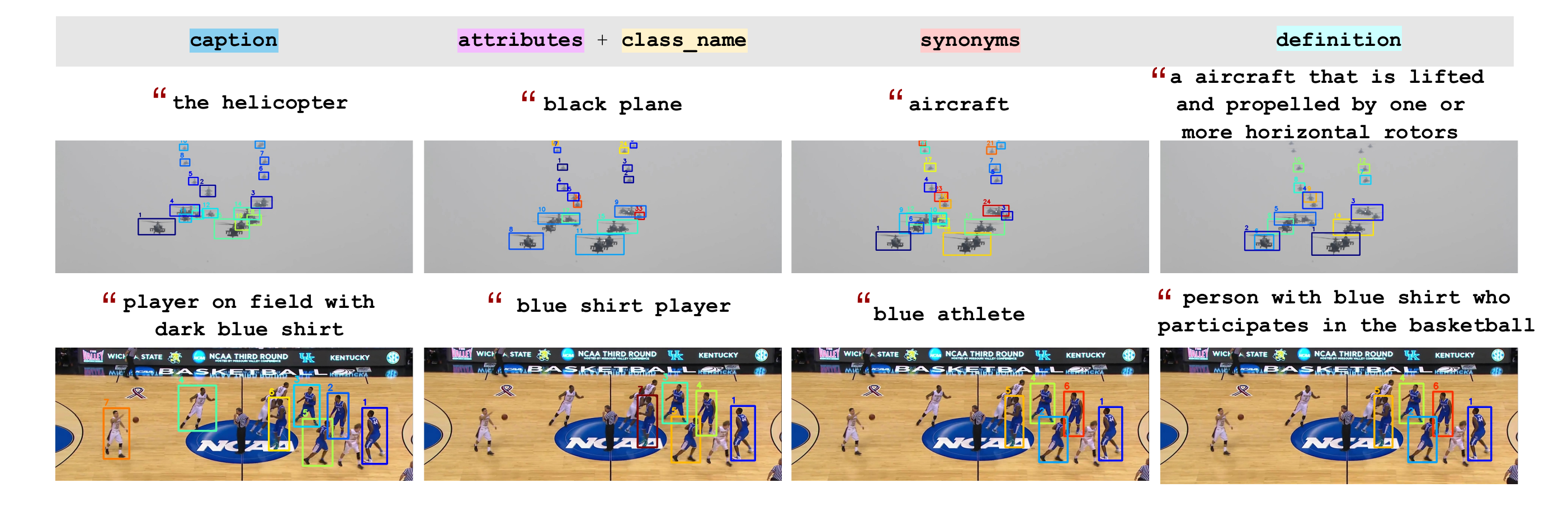}
\caption{KAM-SORT's tracking performance on the G$^2$MOT dataset under different textual description settings. From left to right: detected boxes and tracking IDs when using  \colorbox{caption}{\texttt{caption}} (default setting), \texttt{\colorbox{attribute}{attribution} + \colorbox{classname}{class\_name}}, \colorbox{synonym}{\texttt{synonyms}} (randomly select one synonym from a list of synonyms), \colorbox{definition}{\texttt{definition}}.}
    \label{fig:vis_settings}
\end{figure*}

We conducted four ablation studies as follows:

The first ablation study, presented in Table~\ref{tb:various_settings}, evaluates KAM-SORT's tracking performance on the G$^2$MOT dataset under different annotation settings. These settings include using both \colorbox{attribute}{\texttt{attribute}} and \colorbox{classname}{\texttt{class\_name}}, using \colorbox{definition}{\texttt{definition}}, and using \colorbox{caption}{\texttt{caption}} as the default setting. This study highlights the utility and accuracy of our fine-grained and informative annotations, valuable not only for object tracking but also for many other applications and further exploration. Fig.\ref{fig:vis_settings} illustrates some visualization of tracking performance by KAM-SORT on different annotation settings. Although we do not report tracking performance using \texttt{synonym} in Table~\ref{tb:various_settings} due to it represents a list of synonyms, the illustration in Fig. \ref{fig:vis_settings} demonstrates promising results, indicating the potential for further exploration in the future.

The second ablation study, shown in Table~\ref{tb:Abl_MOT20}, assesses KAM-SORT's performance on the MOT task by comparing it on the MOT20 dataset \cite{dendorfer2020mot20}. To ensure fairness, we disable certain ad-hoc settings, such as employing varying thresholds for individual sequences, an offline interpolation procedure, and internal weights for object detection. YOLOX object detector is used for all trackers to demonstrate the effectiveness of KAM-SORT.
 
The third ablation study, presented in Table~\ref{tab:abl-kamsort}, evaluates two novel components of KAM-SORT: (a) Appearance-Motion Balance, which balances visual appearance and motion cues (Eq. \ref{eq:cost_1}), and (b) Kalman++ (Algorithm \ref{alg:cap}), an alternative algorithm replacing traditional Kalman. The ablation study (\texttt{\#1} v.s. \texttt{\#2}) and (\texttt{\#3}v.s. \texttt{\#4}) shows the importance impact of Kaman++ whereas (\texttt{\#1} v.s. \texttt{\#3}) and (\texttt{\#2} v.s. \texttt{\#4}) shows the importance impact Appearance-Motion Balance. 

The fourth ablation study, shown in Table~\ref{tab:ablation_all}, report KAM-SORT's tracking performance on various hyper-parameters, i.e. vector similarity $\theta$, defined in Eq. \ref{eq:Wa} and uncertainly revision $\alpha$, defined in Algorithm \ref{alg:cap}. 

\vspace{-5mm}
\vspace{-5mm}

\section{Conclusion and Discussion}

In this paper, we introduced a novel generic multi-object tracking (GMOT) framework called \textbf{Grounded-GMOT}, which leverages natural language descriptions to track multiple generic objects in videos. Alongside this framework, we unveiled the G$^2$MOT dataset, which offers diverse object categories, substantial data size, and challenging properties such as a large number of objects, fast motion, significant occlusion, high appearance similarity, and high object density. The \textbf{G$^2$MOT} dataset is annotated with fine-grained language descriptions, including \texttt{synonyms, descriptions, attributes, definitions,} and \texttt{captions}.
Additionally, we presented \textbf{KAM-SORT}, an innovative object association method that incorporates Kaman++, an enhancement of the Kalman Filter, effectively balancing motion and appearance cues. Our extensive experiments demonstrated the remarkable efficacy of the Grounded-GMOT framework in the GMOT task, significantly outperforming existing SOTA OS-GMOT methods. Furthermore, our experiments and ablation studies highlighted KAM-SORT's superior performance compared to all SOTA trackers in both GMOT and MOT tasks.

\vspace{1mm}
\noindent
\textbf{Discussion:} 
In our Grounded-GMOT framework, we utilize Grounding DINO as our preferred VLM for detecting object bounding boxes, using textual description {\texttt{captions}} as the query input as default. However, it is important to recognize the rich diversity of VLMs available in the field, which opens up exciting avenues for deeper exploration. Researchers and users have the opportunity to explore many other alternative VLMs specifically designed for object detection, including noteworthy options like ViLD, RegionCLIP, GLIP, X-DETR, and OWL-ViT (Section \ref{sec:VLM}). Moreover, in our ablation study, we implemented Grounded-GMOT using various textual description settings including \texttt{definitions}, \texttt{synonym}, \texttt{attribute + class\_name}, in addition to the default \texttt{caption}. This showcases the informative annotation of our proposed fine-grained G$^2$MOT dataset, which holds potential for various research and future exploration endeavors. Exploring these additional aspects of G$^2$MOT could lead to enhanced object tracking capabilities and advancements in fields such as surveillance, robotics, and animal welfare.

\newpage

\bibliographystyle{splncs04}
\bibliography{main}

\definecolor{airforceblue}{rgb}{0.36, 0.54, 0.66}
\definecolor{asparagus}{rgb}{0.94, 0.87, 0.8}
\definecolor{almond}{rgb}{0.53, 0.66, 0.42}
\definecolor{babyblue}{rgb}{0.54, 0.81, 0.94}
\definecolor{aureolin}{rgb}{0.99, 0.93, 0.0}

\definecolor{classname}{rgb}{0.98, 0.92, 0.84}
\definecolor{type}{rgb}{0.7, 0.75, 0.71}
\definecolor{synonym}{rgb}{0.96, 0.76, 0.76}
\definecolor{caption}{rgb}{0.54, 0.81, 0.94}
\definecolor{definition}{rgb}{0.67, 0.9, 0.93}
\definecolor{attribute}{rgb}{0.96, 0.73, 1.0}
\definecolor{track_path}{rgb}{1.0, 0.33, 0.64}

\definecolor{airforceblue}{rgb}{0.36, 0.54, 0.66}
\definecolor{asparagus}{rgb}{0.94, 0.87, 0.8}
\definecolor{almond}{rgb}{0.53, 0.66, 0.42}
\definecolor{babyblue}{rgb}{0.54, 0.81, 0.94}
\definecolor{aureolin}{rgb}{0.99, 0.93, 0.0}

\definecolor{classname}{rgb}{0.98, 0.92, 0.84}
\definecolor{type}{rgb}{0.7, 0.75, 0.71}
\definecolor{synonym}{rgb}{0.96, 0.76, 0.76}
\definecolor{caption}{rgb}{0.54, 0.81, 0.94}
\definecolor{definition}{rgb}{0.67, 0.9, 0.93}
\definecolor{attribute}{rgb}{0.96, 0.73, 1.0}

\renewcommand{\thetable}{\Roman{table}}
\renewcommand{\thefigure}{\Roman{figure}}
\setcounter{table}{0}
\setcounter{figure}{0}

\colorlet{punct}{red!60!black}
\definecolor{background}{HTML}{EEEEEE}
\definecolor{delim}{RGB}{20,105,176}
\colorlet{numb}{magenta!60!black}

\lstdefinelanguage{json}{
    basicstyle=\footnotesize\ttfamily,
    numbers=left,
    numberstyle=\scriptsize,
    stepnumber=1,
    numbersep=8pt,
    showstringspaces=false,
    breaklines=true,
    frame=lines,
    backgroundcolor=\color{background},
    literate=
     *{0}{{{\color{numb}0}}}{1}
      {1}{{{\color{numb}1}}}{1}
      {2}{{{\color{numb}2}}}{1}
      {3}{{{\color{numb}3}}}{1}
      {4}{{{\color{numb}4}}}{1}
      {5}{{{\color{numb}5}}}{1}
      {6}{{{\color{numb}6}}}{1}
      {7}{{{\color{numb}7}}}{1}
      {8}{{{\color{numb}8}}}{1}
      {9}{{{\color{numb}9}}}{1}
      {:}{{{\color{punct}{:}}}}{1}
      {,}{{{\color{punct}{,}}}}{1}
      {\{}{{{\color{delim}{\{}}}}{1}
      {\}}{{{\color{delim}{\}}}}}{1}
      {[}{{{\color{delim}{[}}}}{1}
      {]}{{{\color{delim}{]}}}}{1},
}

\newpage
\appendix

\setlength{\abovedisplayskip}{0pt}
\setlength{\belowdisplayskip}{0pt}

In this supplementary, we will further explore the following aspects:
\begin{itemize}
    \item Appendix~\ref{sec:data} provides additional details about G$^2$MOT dataset.
    \item Appendix~\ref{sec:exp} presents further discussion on the proposed Kalman++ and further qualitative comparison between our KAM-SORT with SORT, OC-SORT and DeepOCSORT. 
\end{itemize}

\begin{center}
We will release our dataset and code at:

\url{https://UARK-AICV.github.io/G2MOT}  
\end{center}

\section{G$^2$MOT Dataset}
\label{sec:data}

\subsection{Data Property and Criteria Definition}

GMOT is designed for tracking multiple generic objects. Therefore, our aim is to construct a large-scale dataset that not only features high similarity in appearance (i.e., generic obejcts) but also encompasses challenges such as a large number of objects per frame, high density, substantial occlusion, and fast motion. 

While the the main paper presents comprehensive numerical information for each criteria, this supplementary document examines the diversity of each criterion within every dataset, as illustrated in Table \ref{tb:dataset_comparison}. The diversity thresholds for each criterion have been determined based on our analysis and observation.

\begin{table}[!h]
\vspace{-5mm}
\caption{\textbf{Diversity comparison} of existing \colorbox{almond!50}{MOT} and \colorbox{aureolin!50}{GMOT} datasets. The diversity of each criterion is determined by the thresholds indicated below each criterion.} 
\setlength{\tabcolsep}{3pt}
\renewcommand{\arraystretch}{1.0}
\resizebox{\linewidth}{!}{
\begin{tabular}{l|lc|lllll|ccccc}
\toprule
\multirow{2}{*}{\textbf{Datasets}}   & \multirow{2}{*}{\textbf{Task}} & \multirow{2}{*}{\textbf{NLP}} & \multicolumn{5}{c|}{\textbf{Statistical Information}}& \multicolumn{5}{c}{\textbf{Data Properties}} \\ \cline{4-13}
 & & & \#\textbf{Cat.} & \#\textbf{Vid.} & \#\textbf{Frames} & \#\textbf{Tracks} & \#\textbf{Boxs} & \textbf{Obj.} & \textbf{App.} & \textbf{Den.} & \textbf{Occ.} & \textbf{Mot.}\\  & & & {>10} & {>100} & {>100K} & {>5K} & {>1M} & {>10(>5)} & {>70(>5)} & {>2.5(>0.5))} & {>15(>5)} & {>80(>5)}\\ 
\midrule
\rowcolor{airforceblue!20}
\rowcolor{almond!50}
MOT17~\cite{milan2016mot16}       & MOT  &  \xmark   &      \xmark        &    \xmark      &    \xmark      &    \xmark      &   \xmark   & \textcolor{blue}{\cmark}  & \xmark   & \textcolor{blue}{\cmark} & \xmark   & \textcolor{blue}{\cmark}\\ 
\rowcolor{almond!50}
MOT20~\cite{dendorfer2020mot20}        & MOT  &  \xmark    &        \xmark      &     \xmark     &     \xmark     &     \xmark    &     \textcolor{blue}{\cmark}  & \textcolor{blue}{\cmark} & \xmark & \textcolor{blue}{\cmark} & \textcolor{blue}{\cmark} & \xmark \\ 
\rowcolor{almond!50}
Omni-MOT~\cite{sun2020simultaneous}     & MOT  &  \xmark    &          \xmark    &     --     &      \textcolor{blue}{\cmark}   &    \textcolor{blue}{\cmark}      &    \xmark  & -- & -- & --  & -- & -- \\  
\rowcolor{almond!50}
DanceTrack~\cite{sun2022dancetrack}   & MOT  &  \xmark    &     \xmark        &    \textcolor{blue}{\cmark}      &     \textcolor{blue}{\cmark}     &    \xmark      &  --   & \xmark  & \textcolor{blue}{\cmark} &  \textcolor{blue}{\cmark} & \textcolor{blue}{\cmark} & \textcolor{blue}{\cmark}  \\  
\rowcolor{almond!50}
TAO~\cite{dave2020tao}     & MOT  &  \xmark    &       \textcolor{blue}{\cmark}       &     \textcolor{blue}{\cmark}     &    \textcolor{blue}{\cmark}      &    \textcolor{blue}{\cmark}      &    \xmark  &  \xmark &  \xmark &  \xmark  &  \xmark &  \xmark \\
\rowcolor{almond!50}
SportsMOT~\cite{cui2023sportsmot}     & MOT  &  \xmark    &       \xmark      &    \textcolor{blue}{\cmark}      &    \textcolor{blue}{\cmark}     &      \xmark     &     \xmark  &  \xmark  & \textcolor{blue}{\cmark} &  \xmark & \textcolor{blue}{\cmark} & \textcolor{blue}{\cmark} \\ 

\rowcolor{aureolin!50}
AnimalTrack~\cite{zhang2022animaltrack}  & GMOT &  \xmark   &      \xmark        &      \xmark    &    \xmark       &    \xmark       &     \xmark  & \textcolor{blue}{\cmark}  & \textcolor{blue}{\cmark}  & \textcolor{blue}{\cmark} & \textcolor{blue}{\cmark} & \textcolor{blue}{\cmark}\\ 
\rowcolor{aureolin!50}
GMOT-40~\cite{bai2021gmot}      & GMOT &  \xmark   &         \xmark      &     \xmark      &     \xmark      &    \xmark       &   \xmark    & \textcolor{blue}{\cmark}  & \textcolor{blue}{\cmark} & \textcolor{blue}{\cmark} &  \xmark &  \xmark \\ \midrule
\rowcolor{almond!50}
Refer-KITTI~\cite{wu2023referring}  &  MOT    &  {coarse}   &      \xmark         &      \xmark     &     \xmark      &    \xmark	&  \xmark   &  \xmark &  \xmark &  \xmark  &  \xmark &  \xmark \\ 
\rowcolor{aureolin!50}
\textbf{G$^2$MOT (Ours)}        & GMOT &   {fine}   &     \textcolor{blue}{\cmark}         &   \textcolor{blue}{\cmark}       &   \textcolor{blue}{\cmark}       &   \textcolor{blue}{\cmark}   &  \textcolor{blue}{\cmark}  & \textcolor{blue}{\cmark}  &  \textcolor{blue}{\cmark} & \textcolor{blue}{\cmark} & \textcolor{blue}{\cmark} & \textcolor{blue}{\cmark} \\  \bottomrule
\end{tabular}}
\label{tb:dataset_comparison}
\end{table}

These criteria are defined as follows:

\begin{itemize}
    \item \textbf{Obj.}: average number of objects per frame. \\
    Obj. = $\frac{1}{N}\sum_{t=1}^N{M^t}$, where $M^t$ is the number of objects of frame $t^{th}$ and $N$ is the number of frames.
    
    \item \textbf{Occ.(\%)}: occlusion between objects in a frame, represented by the average ratio of IoU of the bounding boxes in the same frame: \\
    Occ = $\frac{1}{N}\sum_{t=1}^N {\left[\frac{2}{M^{t} \times (M^{t}-1)} \sum_{i=1}^{M^{t}-1} \sum_{j=i+1}^{M^t}{IoU(O_i^t, O_j^t)} \right]}$, where $O_i^t, O_j^t$ are two objects $i^{th}$ and $j^{th}$ in the frame $t^{th}$.
   
    \item \textbf{App.(\%)}: appearance similarity between objects in a frame, calculated by the average cosine similarity of objects in the same frame.  \\ 
    App = $\frac{1}{N}\sum_{t=1}^N{\left[\frac{2}{M^t \times (M^t-1)} \sum_{i=1}^{M^{t}-1} \sum_{j=i+1}^{M^t}{cos<F(O_i^t), F(O_j^t)>}\right]}$
    , where $F(O_i^t), F(O_j^t)$ are feature embeddings of two objects $i^{th}$ and $j^{th}$ in the frame $t^{th}$.
  
    \item \textbf{Den.}: density of objects in a frame, computed by the maximum number of objects at the same pixel. \\
    Den. = $\frac{1}{N}\sum_{t=1}^N{Max(H^t)}$, where $H^t$ is a density heatmap of the frame $t^{th}$, with each element $H^t[i,j]$ represents the number of objects occupying the pixel $[i,j]$. $Max(.)$ returns the maximum value.
    
    \item \textbf{Mot.(\%)}: motion speed of objects in a video, calculated by the average ratio of the IoU of the bounding boxes in the same track in consecutive frames. \\
    Mot. = $\frac{1}{\sum_{i=1}^{K}[Length(Track_i) - 1]}{\sum_{i=1}^{K} \sum_{t=1}^{|Length(Track_i)| - 1} IoU(Track_i^t, Track_i^{t+1})}$, \\
    where K is the number of tracks in the video, and $Track_i^t$ is a track corresponding to a track ID $i^{th}$ throughout the video at frame $t^{th}$.

\end{itemize}

\subsection{Annotation}
\vspace{-2mm}
\label{sec:data_structure}
The annotation format of JSON files in our G$^2$MOT dataset is as follows:
\renewcommand{\ttdefault}{pcr}
\noindent

\begin{minipage}{.85\textwidth}
\centering
\begin{lstlisting}[language=json, caption={ JSON format defined in G$^2$MOT dataset. }, label={lst:anno_fmt},  basicstyle=\ttfamily\scriptsize, columns=fixed, floatplacement=htbp]
"video": [{
    "id": int,
    "video_path": str
}],
"tracking_query": [{
    "id": int,
    "video_id"; int,
    "is_eval": bool,
    "type": str, #"superset" or "subset"
    "superset_idx": int #-1 if type "superset", 
    "class_name": str,
    "synonyms": [str],
    "definition": str,
    "attributes": [str],
    "track_path": str,
    "caption": str,
}]
\end{lstlisting}
\end{minipage}

\begin{figure*}[t]
    \centering
    \includegraphics[width=\textwidth]{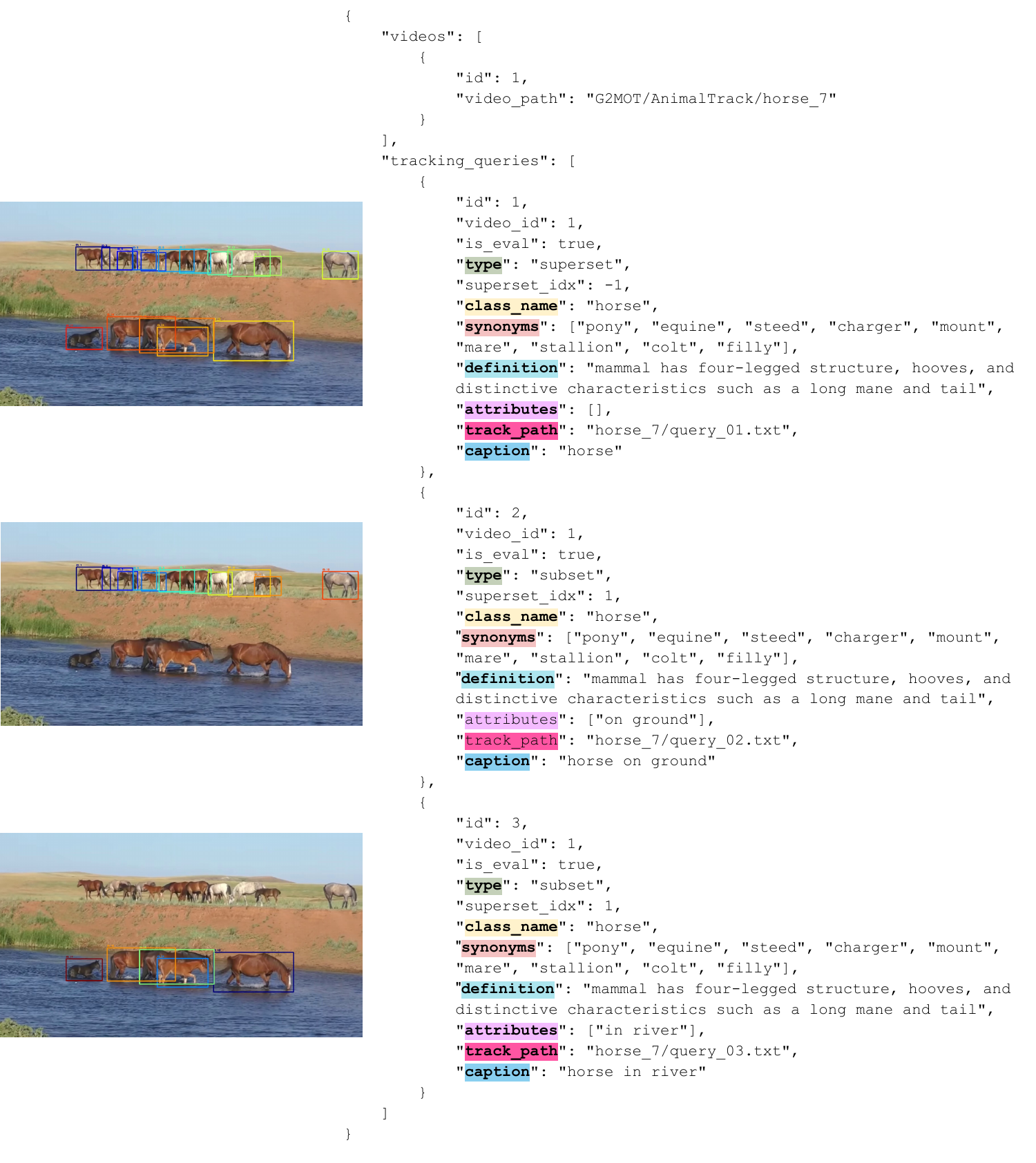}
    \caption{As an illustration of our annotation, consider the video ``horse\_7''. It includes one superset representing all horses within the scene and two subsets that specifically identify horses on the ground and horses in the river.}
    \label{fig:data_annotation}
\end{figure*}

\begin{figure*}[t]
    \centering
    \includegraphics[width=0.9\linewidth]{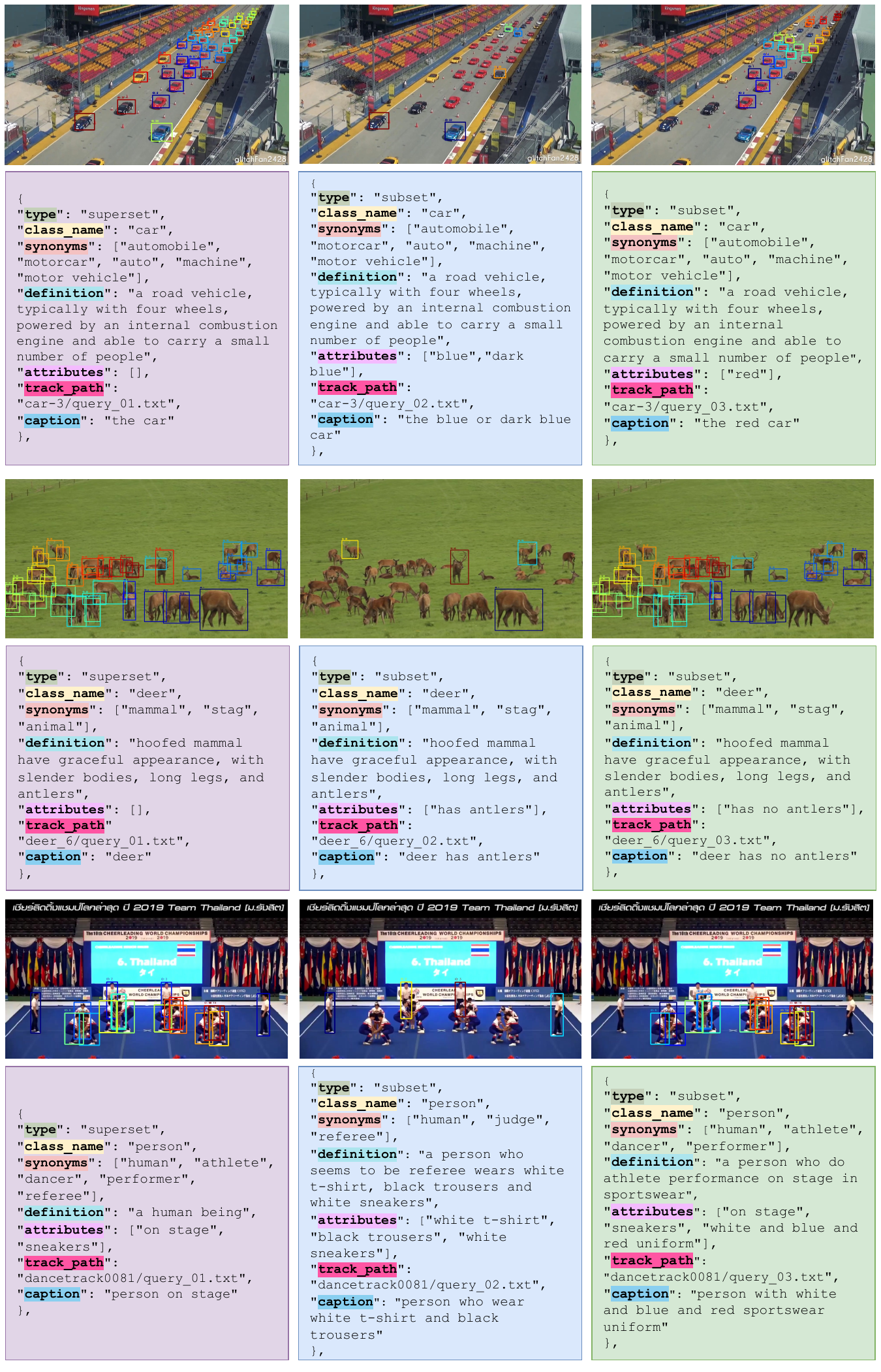}
    \caption{Additional examples of textual description provided in our G$^2$MOT.}
    \label{fig:fig_supp_w_anno}
\end{figure*}

A complete annotation of a video is depicted in Figure \ref{fig:data_annotation}. Additionally, Figure \ref{fig:fig_supp_w_anno} provides textual descriptions for additional videos.

\subsection{Statistical Information of G$^2$MOT}
\label{sec:statis}
Table \ref{tab:data_for_each_class} provides the statistical information for each class object in G$^2$MOT dataset.

\begin{figure*}[t]
    \centering
    \includegraphics[width=0.8\textwidth]{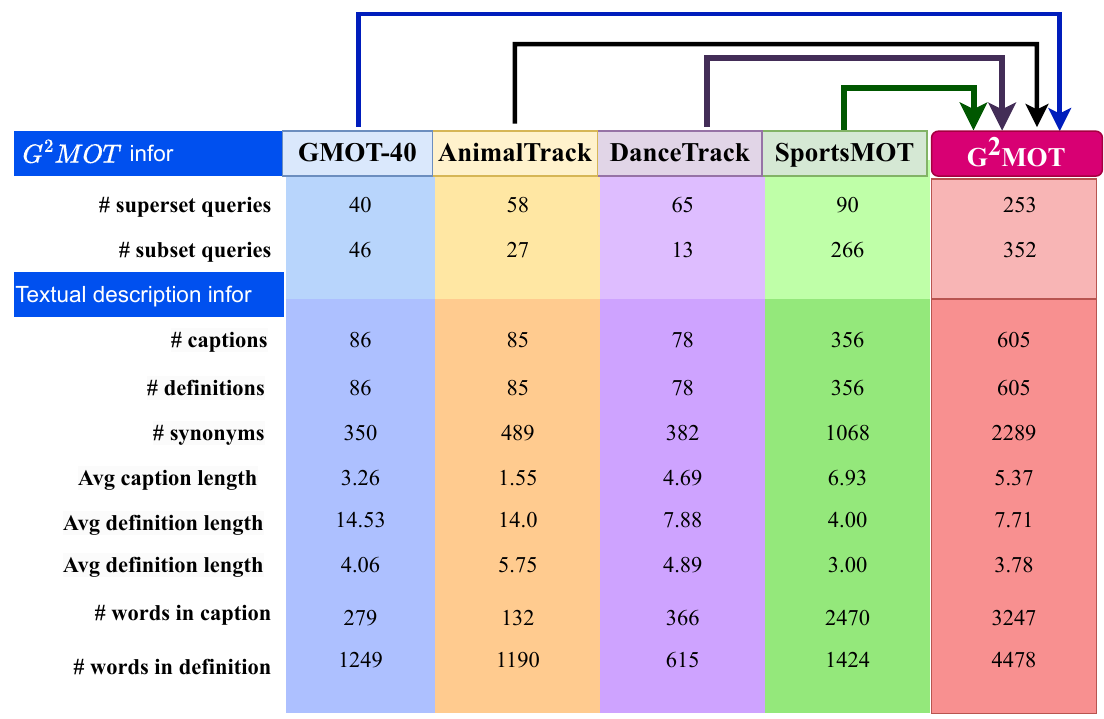}
    \caption{Statistical information from each sub-dataset (GMOT-40, AnimalTrack, DanceTrack, SportMOT) for the construction of our G$^2$MOT dataset.}
    \label{fig:all_infos}
\end{figure*}

G$^2$MOT integrates datasets from GMOT-40, AnimalTrack, DanceTrack, and SportMOT, thereby presenting a diverse array of challenges for GMOT as outlined in Table \ref{tb:dataset_comparison}. The statistical details from each sub-dataset are provided in \cref{fig:all_infos}. Specifically, SportMOT emphasizes tracking across extended intervals and distinguishing between similar appearances. This task is complicated by factors such as uniform attire and the presence of small, distant objects. Notably, SportMOT exhibits the highest subset-to-superset ratio, making object localization challenging due to subtle visual discrepancies.

In addition, both SportMOT and DanceTrack datasets highlight significant movement and substantial occlusion, along with instances of high similarity in appearance. Meanwhile, GMOT-40 and AnimalTrack datasets present a multitude of small, similarly appearing objects with high density, as well as a high average object count per frame.

\begin{table}[t]
\centering
\setlength{\tabcolsep}{8pt}
\renewcommand{\arraystretch}{0.9}
    \centering
    \caption{Statistics information of G$2$MOT. \# denotes the quantity of the respective items.}
    \resizebox{0.85\linewidth}{!}{\begin{tabular}{lrrrrr}
        \toprule
        \textbf{Class Name} & \# \textbf{Frames} & \# \textbf{Objects} & \# \textbf{Boxes} & \# \textbf{superset} & \# \textbf{subset}\\
        \midrule
        
        airplane & 1181 & 81 & 23307 & 4 & 0\\
        athlete & 90642 & 1282 & 591146 & 45 & 131 \\
        balloon & 3218 & 557 & 83430 & 4  & 11\\
        ball & 729 & 181 & 20399 & 4 & 8 \\
        bird & 3462 & 307 & 70972 & 5 & 4\\
        boat & 1472 & 143 & 29491 & 4 & 5 \\
        car & 2023 & 221 & 33893 & 4 & 7 \\
        chicken & 7615 & 226 & 54710 & 5 & 19 \\
        deer & 2762 & 222 & 53339 & 7 & 2\\
        dolphin & 1718 & 167 & 31112 & 6 & 0\\
        duck & 4851 & 277 & 79416 & 6 & 0\\
        fish & 569 & 291 & 21922 & 4 & 2 \\
        goose & 1773 & 195 & 33120 & 5 & 0\\
        horse & 6556 & 281 & 69524 & 7 & 4\\
        insect & 659 & 297 & 14107 & 4 & 0 \\
        penguin & 1844 & 137 & 30312 & 6 & 0 \\
        person & 83623 & 939 & 688535 & 69 & 22 \\
        pig & 1531 & 151 & 32273 & 5 & 0 \\
        player & 85474 & 1254 & 615717 & 45 & 135 \\
        rabbit & 1558 & 313 & 33961 & 5 & 0 \\
        stock & 1128 & 143 & 34058 & 4 & 2 \\
        zebra & 1378 & 99 & 22331 & 5 & 0 \\
        \bottomrule
    \end{tabular}}
    \label{tab:data_for_each_class}
\end{table}

\section{Kalman++ Discussion and Qualitative Results}
\label{sec:exp}
\subsection{Kalman++ Discussion }
Object tracking is a challenging task, especially under conditions such as low frame rates, camera movement, object deformation, etc which introduce noise to estimators. Linear estimators like the Kalman Filter are particularly affected. While the Kalman Filter assesses the state uncertainty covariance, many Kalman Filter-based trackers primarily rely on their own predictions of the next state in the future frame during the object association phase. In this process, the uncertainty covariance is overlooked and typically only involved in calculating the Kalman Gain. To address this, we introduce a second-phase matching into Kalman, creating Kalman++. This enables effective utilization of uncertainty covariance while mitigating the negative effects of the aformentioned conditions.

\subsection{Qualitative Results}

In addition to Figure 4 in the main paper, which primarily addresses \textit{substantial occlusion} issues in tracking, we provide further illustrations demonstrating the effectiveness of KAM-SORT as follows:

\vspace{3mm}
\noindent
\textbf{Qualitative results on fast motion}

\begin{figure*}[h]
    \centering
    \includegraphics[width=\linewidth]{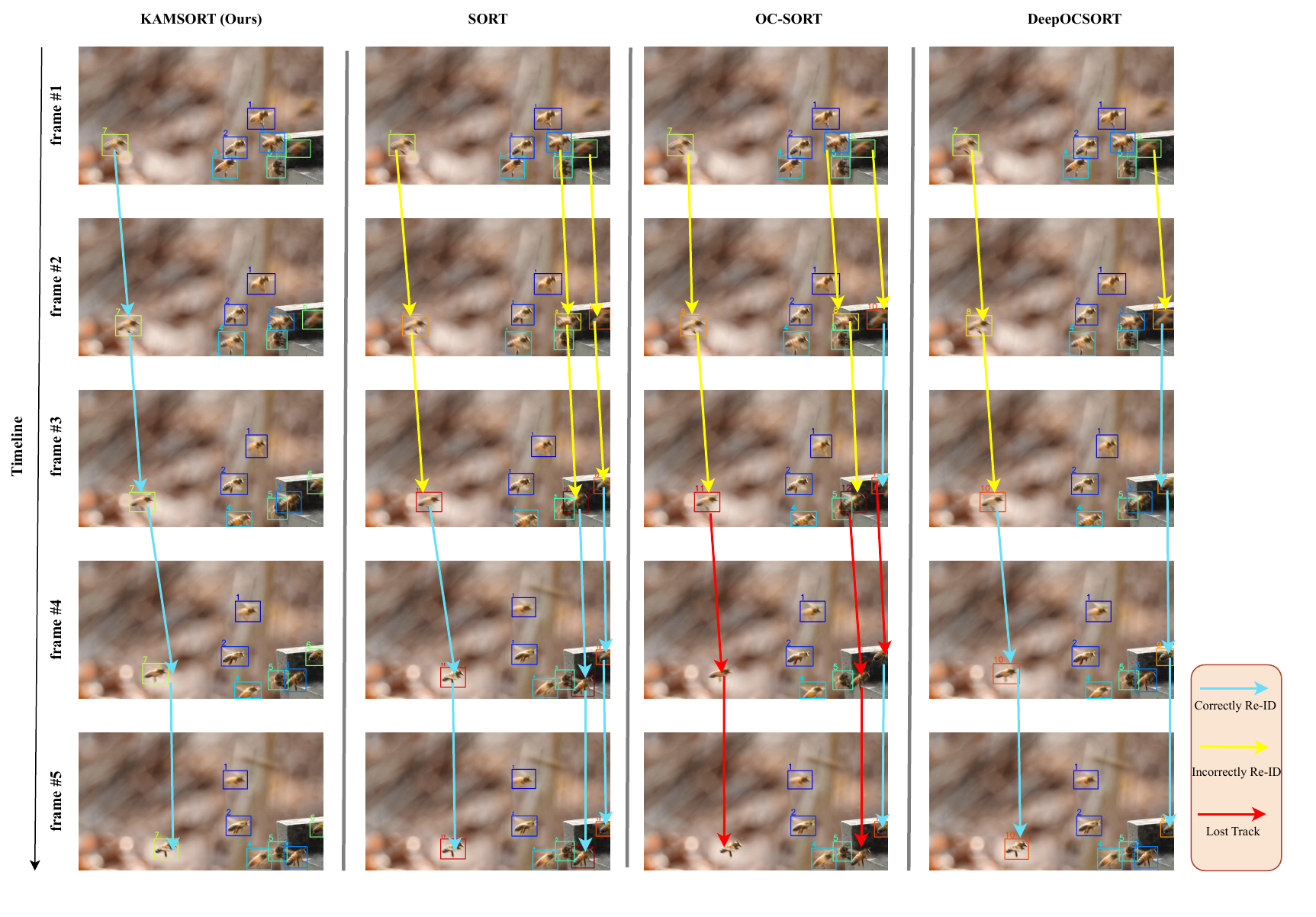}
    \caption{Tracking comparison on \textit{fast motion} objects between our KAM-SORT with SORT, OC-SORT and DeepOCSORT on the video ``insect-3''.}
    \label{fig:compare_fastmoving}
\end{figure*}

\cref{fig:compare_fastmoving} illustrates a qualitative comparison of \textit{fast-moving object tracking }between our KAM-SORT and SORT, OC-SORT, and DeepOCSORT. In this scenario, the tracked objects (i.e., bees) exhibit high-speed movement, leading to blurring effects that impact both their appearance and detected bounding boxes. Consequently, issues such as lost track and incorrectly-ReID are common in SORT, OC-SORT, and DeepOCSORT. 



\vspace{3mm}
\noindent
\textbf{Qualitative results on high similarity in appearance}

\begin{figure*}[h]
    \centering
    \includegraphics[width=\linewidth]{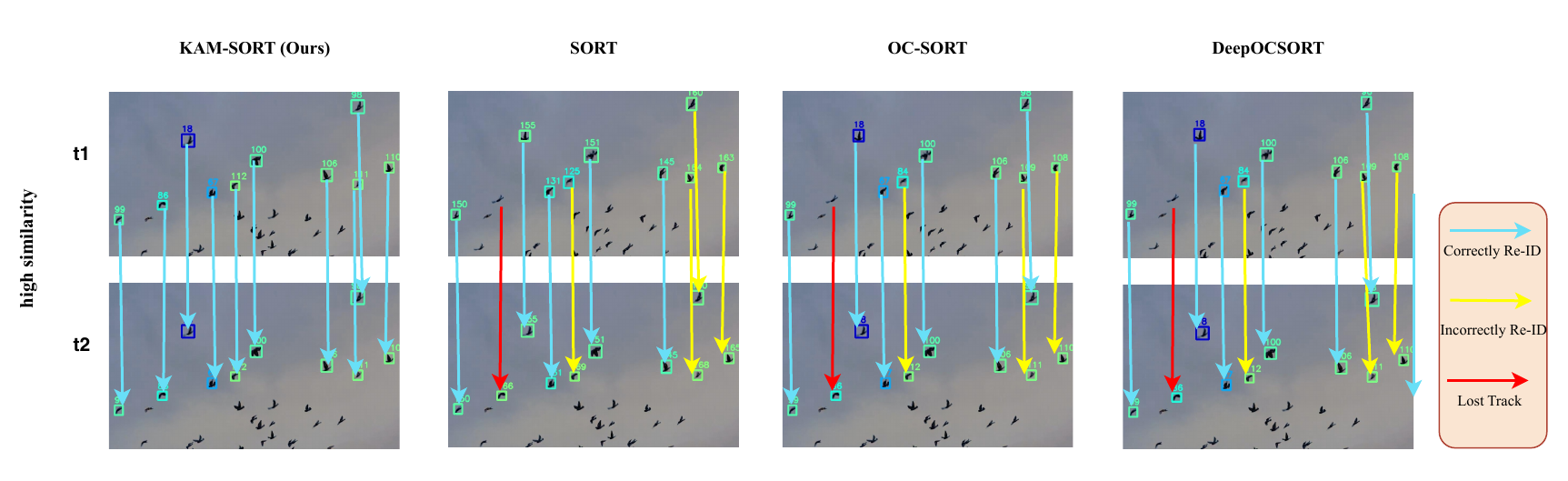}
    \caption{Tracking comparison on \textit{high similarity in appearance} objects between our KAM-SORT with SORT, OC-SORT and DeepOCSORT on the video ``bird-0''.}
    \label{fig:fig_appearance}
\end{figure*}

\cref{fig:fig_appearance} illustrates a qualitative comparison of \textit{nearly identical objects} between our KAM-SORT and SORT, OC-SORT, and DeepOCSORT. In this scenario, the tracked objects (i.e., birds) exhibit highly similar appearances. Consequently, incorrectly re-ID is frequent in SORT, OC-SORT, and DeepOCSORT. 

\vspace{3mm}
\noindent
\textbf{Qualitative results on camera motion}

\begin{figure*}[h]
    \centering
    \includegraphics[width=\linewidth]{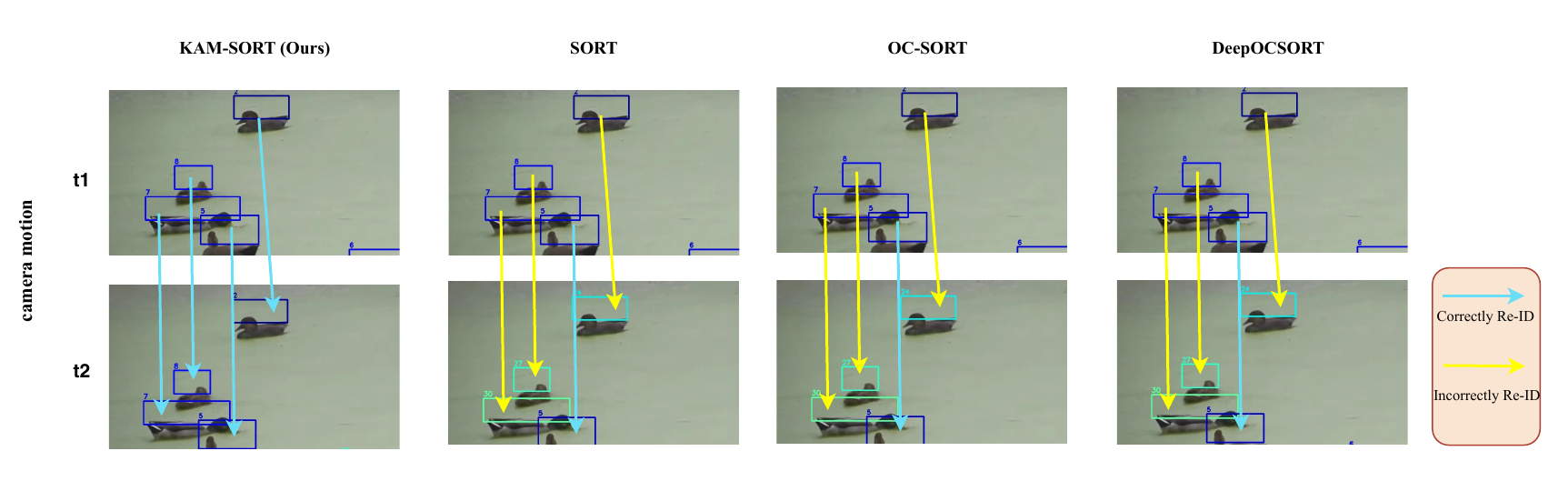}
    \caption{Tracking comparison when both camera and object moving between our KAM-SORT with SORT, OC-SORT and DeepOCSORT on the video ``duck-4''.}
    \label{fig:fig_camera_motion}
\end{figure*}

\cref{fig:fig_camera_motion} illustrates a qualitative comparison of \textit{tracking objects with camera motion} between our KAM-SORT and SORT, OC-SORT, and DeepOCSORT. In this scenario, both tracked objects (i.e., ducks) and cameras moving, which results in a big off-set between the predicted boxes and the actual bounding boxes around objects. Consequently, incorrectly re-ID is frequent in SORT, OC-SORT, and DeepOCSORT. 

By flexibly balancing between appearance and motion cues, our KAM-SORT enable robust maintenance of object IDs throughout the tracking process across various conditions.

\end{document}